\documentclass[sn-basic]{sn-jnl}% Basic Springer Nature Reference Style/Chemistry Reference Style
\jyear{2021}%

\theoremstyle{thmstyleone}%
\newtheorem{theorem}{Theorem}%  meant for continuous numbers
\newtheorem{proposition}[theorem]{Proposition}% 

\theoremstyle{thmstyletwo}%
\newtheorem{example}{Example}%
\newtheorem{remark}{Remark}%

\theoremstyle{thmstylethree}%
\newtheorem{definition}{Definition}%

\usepackage{makecell}
\usepackage{multirow}
\usepackage{array}
\usepackage{colortbl}
\usepackage{booktabs}
\usepackage{dashrule}
\usepackage{arydshln}
\usepackage{verbatim}
\usepackage{float}
\usepackage{subfig}
\usepackage{soul, color, xcolor}
\usepackage{pifont}

\soulregister{\citep}7
\soulregister{\citet}7
\soulregister{\ref}7
\soulregister{\Cref}7

\newcommand{\NUMDATASET}{11}
\newcommand{\NUMTURN}{471k}

\newcommand{\NUMDomainSchema}{336}

\newcommand*{\ours}{\textsc{ESAinsTOD}}
\newcommand*{\llamaSmall}{\textsc{Llama 2 7B}}
\newcommand*{\llama}{\textsc{Llama}}

\newcommand{\setTextBG}[2]{\sethlcolor{#1}\hl{#2}}

\newcommand{\hlyellow}[1]{#1}
\newcommand{\sechl}[1]{#1}

\begin{document}

\title[A Unified End-to-End Schema-Aware Task-Oriented Dialog Modeling]{\ours{}: A Unified End-to-End Schema-Aware Instruction-Tuning Framework for Task-Oriented Dialog Modeling}


\author[1]{\fnm{Dechuan} \sur{Teng}}\email{dcteng@ir.hit.edu.cn}
\author[2]{\fnm{Chunlin} \sur{Lu}}\email{234711290@csu.edu.cn}

\author*[2]{\fnm{Libo} \sur{Qin}}\email{lbqin@csu.edu.cn}
\author*[1]{\fnm{Wanxiang} \sur{Che}}\email{car@ir.hit.edu.cn}

\affil[1]{\orgname{Research Center for Social Computing and Information Retrieval, Harbin Institute of Technology}, \state{Harbin}, \country{China}}
\affil[2]{\orgname{School of Computer Science and Engineering, Central South University}, \state{Changsha}, \country{China}}

%\author*[1,2]{\fnm{First} \sur{Author}}\email{iauthor@gmail.com}
%\affil*[1]{\orgdiv{Department}, \orgname{Organization}, \orgaddress{\street{Street}, \city{City}, \postcode{100190}, \state{State}, \country{Country}}}
%
%\affil[2]{\orgdiv{Department}, \orgname{Organization}, \orgaddress{\street{Street}, \city{City}, \postcode{10587}, \state{State}, \country{Country}}}
%
%\affil[3]{\orgdiv{Department}, \orgname{Organization}, \orgaddress{\street{Street}, \city{City}, \postcode{610101}, \state{State}, \country{Country}}}


\abstract{Existing end-to-end modeling methods for modular task-oriented dialog systems are typically tailored to specific datasets, making it challenging to adapt to new dialog scenarios.
In this work, we propose \ours{}, a unified \textit{\textbf{E}}nd-to-end \textit{\textbf{S}}chema-\textit{\textbf{A}}ware \textit{\textbf{Ins}}truction-tuning framework for general \textit{\textbf{T}}ask-\textit{\textbf{O}}riented \textit{\textbf{D}}ialog modeling.
% This framework leverages Large Language Models (LLMs) and demonstrates flexible adaptation to various dialogue task flows and schemas.
\hlyellow{This framework introduces a structured methodology to go beyond simply fine-tuning Large Language Models (LLMs), enabling flexible adaptation to various dialogue task flows and schemas.}
% Specifically, we introduce two alignment mechanisms to make the fine-tuned system  instruction-aware and schema-aware
\hlyellow{Specifically, we leverage full-parameter fine-tuning of LLMs and introduce two alignment mechanisms to make the resulting system both instruction-aware and schema-aware}: (\textit{i}) \textit{instruction alignment}, which ensures that the system faithfully follows task instructions to complete various task flows from heterogeneous TOD datasets; and (\textit{ii}) \textit{schema alignment}, which encourages the system to make predictions adhering to the specified schema.
In addition, we employ \textit{session-level end-to-end modeling}, which allows the system to access the results of previously executed task flows within the dialogue history, to bridge the gap between the instruction-tuning paradigm and the real-world application of TOD systems.
% Empirical results show that 
\hlyellow{Empirical results show that while a fine-tuned LLM serves as a strong baseline, our structured approach provides significant additional benefits.
In particular, our findings indicate that:}
(\textit{i}) \ours{} outperforms state-of-the-art models by a significant margin on end-to-end task-oriented dialog modeling benchmarks: CamRest676, In-Car and MultiWOZ;
% (\textit{ii}) \ours{} exhibits superior generalization capabilities compared to existing models across various low-resource settings, and the proposed alignment mechanisms significantly enhance the zero-shot performance of the dialog system;
(\textit{ii}) \hlyellow{more importantly, it exhibits superior generalization capabilities across various low-resource settings, with the proposed alignment mechanisms significantly enhancing zero-shot performance};
and (\textit{iii}) our instruction-tuning paradigm substantially improves the model's robustness against data noise and cascading errors.}




\maketitle

\section{Introduction}\label{sec:intro}
Task-oriented dialogue (TOD) systems are designed to assist users in accomplishing a wide range of tasks, such as travel planning and restaurant reservation, through natural language communications.
These systems are fundamental in developing various virtual assistants and have recently garnered extensive attention~\citep{DBLP:conf/eacl/Rojas-BarahonaG17,DBLP:conf/acl/HamLJK20,DBLP:conf/aaai/YangLQ21,DBLP:conf/acl/SuSMG0LZ22,DBLP:conf/sigir/HeDYSHSL22}.
A typical TOD system adopts a modularized pipeline architecture~\citep{DBLP:journals/pieee/YoungGTW13,DBLP:journals/ftir/GaoGL19}, where the natural language understanding (NLU) module is responsible for parsing user utterances into semantic frames consisting of intents and slot-value pairs, the dialog management (DM) module for maintaining the dialog state (i.e., user goal), managing database queries, and then deciding next system actions, and the natural language generation (NLG) module for converting the system actions into natural language responses.
With the rapid development of pre-trained language models (PLMs), many methods have been proposed to improve the performance of each module~\citep{DBLP:conf/emnlp/QinCLWL19,DBLP:conf/acl/KimYKL20,DBLP:conf/sigdial/HeckNLGLMG20,DBLP:conf/acl/0088GWYQWHZMCCD22}.
However, the pipeline architecture often suffers from error propagation due to the independent training of each module, where the accumulated errors between modules will lead to the overall performance degradation of the dialog system.
Therefore, much work explores modeling the modularized TOD system in an end-to-end (E2E) trainable manner to alleviate error propagation~\citep{DBLP:conf/nips/Hosseini-AslMWY20, DBLP:conf/aaai/YangLQ21,DBLP:journals/tacl/PengLLSLG21}.
For example, UBAR~\citep{DBLP:conf/aaai/YangLQ21} incorporates dialogue state tracking (DST), policy planning (POL), and natural language generation in a single auto-regressive language model.
Subsequently, dialog pre-training, where vanilla PLMs are pre-trained on large-scale dialog corpora with well-designed dialog objectives, is also proposed to build pre-trained conversation models (PCMs) that enrich conversational knowledge and can be further fine-tuned to improve downstream dialog tasks~\citep{DBLP:journals/corr/abs-2009-13570,DBLP:conf/emnlp/WuHSX20,DBLP:conf/acl/ZhangSGCBGGLD20,DBLP:conf/acl/BaoHWWW20,DBLP:conf/acl/SuSMG0LZ22,DBLP:conf/sigir/HeDYSHSL22}.

\begin{figure*}[t]
    \centering
    \subfloat[Conventional PCMs]{\includegraphics[height=0.50\textwidth]{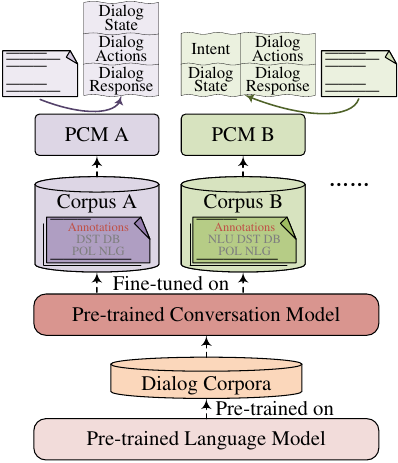}\label{fig:conventional_pcms}}\hspace{0.01\textwidth}
    \subfloat[Our Instruction-tuned LLM for TOD]{\includegraphics[height=0.50\textwidth]{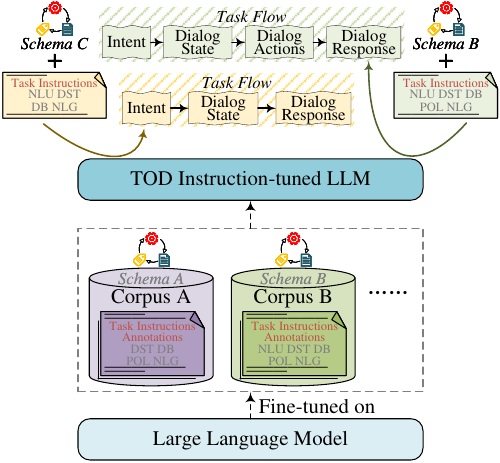}\label{fig:instruction_tuned_llm}}
    \caption{Comparison between conventional PCMs and our instruction-tuned LLM for task-oriented dialog systems. In Figure (a), PCM A and PCM B are derived by fine-tuning pre-trained models on specific corpora A and B, respectively. Each PCM is tailored for a particular workflow and lacks the flexibility to apply to other dialogue scenarios. In contrast, as depicted in Figure (b), the TOD Instruction-tuned LLM is developed by fine-tuning a large language model on multiple corpora, employing both instruction-aware and schema-aware mechanisms. This design enables it to generalize effectively to rare or unseen dialogue scenarios.}
    \label{fig:pcm_comparison}
\end{figure*}

% Despite these remarkable progresses in end-to-end trainable TOD systems, there are still several drawbacks:
\hlyellow{While these methods have advanced the field, the advent of Large Language Models (LLMs) presents new opportunities and challenges.
A straightforward approach is to simply fine-tune a powerful LLM on existing TOD datasets.
However, this often fails to address several persistent drawbacks:}
\begin{itemize}
    % Therefore, the previous work does not consider the correlation between the annotations and the corresponding schema (e.g., database structure and query APIs) in both the pre-training and fine-tuning stages, making it difficult to transfer directly to new dialog scenarios.
    \item \textit{Weak adaptability.} Existing models are tailored to specific datasets covering certain domains and dialog task flows. Moreover, the core idea of existing PCMs is to extract and summarize general dialog knowledge from various TOD datasets during the dialog pre-training stage. \hlyellow{This paradigm does not explicitly account for the tight coupling between data annotations and their corresponding schema (e.g., database structure and available APIs). Consequently, adapting these models to new dialogue scenarios with different underlying schemas remains a significant challenge.} Fig.~\ref{fig:conventional_pcms} illustrates the general process of applying PCMs to the TOD field.
    \item \textit{Insufficient annotation exploitation.} To exploit various public heterogeneous TOD datasets that have incomplete annotations (e.g., data only annotated with DST or POL), existing methods decouple them into different sub-task datasets on which PLMs are pre-trained with multi-task learning objectives~\citep{DBLP:conf/acl/SuSMG0LZ22,DBLP:journals/corr/abs-2312-16864}, without considering the correlation between tasks. Therefore, these resulting PCMs exhibit limited performance on end-to-end TOD modeling.
\end{itemize}

% End-to-end Schema-aware Instruction-tuning for Task-oriented Dialog (ESAinsTOD)
In this work, to address the above two issues, we propose a \textit{\textbf{simple}} yet \textit{\textbf{effective}} instruction-tuning framework, \ours{} (\textit{\textbf{E}}nd-to-end \textit{\textbf{S}}chema-\textit{\textbf{A}}ware \textit{\textbf{Ins}}truction-tuning for general \textit{\textbf{T}}ask-\textit{\textbf{O}}riented \textit{\textbf{D}}ialog modeling).
% As shown in Fig.~\ref{fig:instruction_tuned_llm}, our framework can build a unified TOD system that can generalize to various dialog tasks and scenarios.
\hlyellow{Rather than relying solely on the inherent capabilities of an LLM backbone, our framework introduces a structured approach to unify heterogeneous data and tasks, enabling the model to generalize across diverse dialogue scenarios, as shown in Fig.~\ref{fig:instruction_tuned_llm}.}
Specifically, we introduce \textit{\textbf{instruction alignment}} to link different TOD tasks with corresponding annotations, allowing the model to faithfully follow task instructions to complete given dialog tasks, thus unifying various TOD datasets with heterogeneous annotations.
Since task outputs are related to the corresponding schema, we argue that it is non-trivial to make the model aware of the schema information.
Therefore, we introduce \textit{\textbf{schema alignment}} to force model predictions to be consistent with the specified schema, providing an opportunity to generalize to any dialog scenario with different schemas.
Based on the proposed framework, we construct a multi-turn end-to-end instruction-tuning TOD corpus from \NUMDATASET{} public TOD datasets, which cover a number of domains and scenarios and are partially or fully annotated with intents, dialogue states, database results, and dialogue acts.
The detailed statistics of the corpus are shown in Table~\ref{table:datasets}.
In total, our corpus encompasses over \NUMTURN{} dialogue turns across \NUMDomainSchema{} domain schemas.

% Experimental results show our instruction-tuned model achieves comparable or even better performance on multiple TOD benchmarks, including intent recognition, dialogue state tracking, and end-to-end (E2E) dialogue modeling.
% We also conduct extensive analytical experiments to investigate the advantages of our instruction-tuning paradigm in terms of generalization and error propagation reduction.
\hlyellow{Our experiments are designed to carefully distinguish the performance gains attributable to the powerful LLM backbone from the benefits conferred by our framework itself.
The results demonstrate that while a fine-tuned LLM establishes a strong baseline, our proposed framework (1) significantly enhances generalization capabilities across diverse dialogue scenarios, (2) improves data efficiency in low-resource settings, and (3) effectively mitigates error propagation that is a persistent challenge in both pipeline and end-to-end systems.}
This simple framework can fully consolidate the annotations from various TOD datasets into a single model, efficiently transferring to new dialogue scenarios through instruction and schema alignment. 
Furthermore, schema-aware instruction-tuning establishes clearer boundaries between end-to-end TOD modeling in various scenarios, thereby reducing the frequency of cascading errors by mitigating the interference among training data.

In summary, our contributions are three-fold:
\begin{itemize}
    % which can benefit from the rich knowledge of large language models and various TOD datasets with heterogeneous annotations.
    \item To the best of our knowledge, we are the first to explore a unified instruction-tuning framework for end-to-end task-oriented dialog modeling. \hlyellow{The framework systematically structures heterogeneous dialog data to effectively unlock the generalization capabilities of large language models for complex, multi-domain conversations.}
    \item We construct a multi-turn end-to-end instruction-tuning corpus and obtain an efficient task-oriented dialogue model by fine-tuning an open-source LLM on this corpus. The open corpus covers various dialog scenarios, with each dialog session annotated with specific schema definitions, task instructions, and their corresponding task outputs. 
    This construction paradigm facilitates the establishment of a benchmark for developing and evaluating generalizable TOD systems. The trained model, datasets, and codes are publicly available at \url{https://github.com/AaronTengDeChuan/ESAinsTOD}.
    % We also conduct extensive experiments to verify and analyze the strong generalization ability of \ours{}. We believe that our work will provide a new perspective for the research in generalizable TOD systems.
    \item Results on multiple datasets demonstrate the effectiveness of our proposed task-oriented instruction-tuning framework. \hlyellow{We also conduct extensive experiments to verify and analyze the strong generalization, data efficiency, and error mitigation benefits of \textsc{ESAinsTOD}. We believe that our work will provide a new perspective for building robust and adaptable TOD systems in the era of LLMs.}
\end{itemize}

\begin{table}[t]
% \centering
\begin{center}
\begin{minipage}{0.95\textwidth}
% \normalsize
\setlength{\tabcolsep}{1.0mm}
\caption{Detailed statistics of our end-to-end instruction-tuning TOD corpus}\label{table:datasets}
\begin{tabular*}{\linewidth}{@{\extracolsep{\fill}}lcccccrrcc@{\extracolsep{\fill}}}
\toprule
\makebox[0.170\textwidth][l]{\multirow{2}{*}{\textbf{Dataset}}} & \multicolumn{5}{c}{\textbf{Annotations}} & \makebox[0.09\textwidth][r]{\multirow{2}{*}{\textbf{\texttt{\#} Sess.}}} & \makebox[0.10\textwidth][r]{\multirow{2}{*}{\textbf{\texttt{\#} Turn}}} & \makebox[0.09\textwidth][c]{\multirow{2}{*}{\textbf{\texttt{\#} Dom.}}} & \makebox[0.12\textwidth][c]{\multirow{2}{*}{\textbf{\texttt{\#} Sch.}}} \\ \cmidrule{2-6}
 & {NLU} & {DST} & {DB} & {POL} & {NLG} & & & \\ \midrule
CamRest676 & $\times$ & \checkmark & \checkmark & $\times$ & \checkmark & 676 & 2,744 & 1 & 1 (0) \\
In-Car & $\times$ & \checkmark & \checkmark & $\times$ & \checkmark & 3,028 & 7,876 & 3 & 3 (0) \\
MultiWOZ 2.1 & $\times$ & \checkmark & \checkmark & \checkmark & \checkmark & 10,433 & 71,484 & 7 & 7 (0) \\
SGD & \checkmark & \checkmark & \checkmark & \checkmark & \checkmark & 22,593 & 229,156 & 20 & 270 (528) \\
Frames & $\times$ & \checkmark & $\times$ & $\times$ & \checkmark & 1,367 & 9,579 & 1 & 1 (0) \\
BiToD & \checkmark & \checkmark & \checkmark & \checkmark & \checkmark & 3,689 & 36,231 & 5 & 5 (7) \\
STAR & \checkmark & $\times$ & \checkmark & \checkmark & \checkmark & 5,803 & 53,558 & 13 & 13 (24) \\
BANKING77 & \checkmark & $\times$ & $\times$ & $\times$ & $\times$ & 13,242 & 13,242 & 1 & 1 (0) \\
CLINC150 & \checkmark & $\times$ & $\times$ & $\times$ & $\times$ & 22,500 & 22,500 & 10 & 10 (0) \\
HWU64 & \checkmark & $\times$ & $\times$ & $\times$ & $\times$ & 11,036 & 11,036 & 18 & 18 (0) \\
SNIPS & \checkmark & $\times$ & $\times$ & $\times$ & $\times$ & 14,484 & 14,484 & 7 & 7 (0) \\
\midrule
TOTAL &  &  &  &  &  & 108,851  & 471,890 & 86 & 336 (559) \\
\bottomrule
\end{tabular*}
\footnotetext{These datasets include either subsets or the complete set of annotations pertinent to task-oriented dialogue. In this context, ``DB'' denotes database query results, ``Sess.'' refers to the number of dialogue sessions, and ``Dom.'' indicates the number of domains. In the ``Sch.'' column, the figures outside the parentheses reflect the number of domain schemas, while those inside the parentheses represent the count of intent schemas.}
\end{minipage}
\end{center}
\end{table}

\section{Preliminaries}\label{sec:preliminaries}

In this section, we first formally define the workflow of the traditional modularized TOD system and then present how to model it in an end-to-end manner.
Finally, we introduce the backbone models for our framework, including \llama{}~\citep{DBLP:journals/corr/abs-2307-09288} and Qwen2.5~\citep{DBLP:journals/corr/abs-2412-15115}, open-source decoder-only large language models, which are naturally suitable for end-to-end TOD modeling.

\subsection{Problem definition}
\subsubsection{Notations}\label{sec:notations}
Let $\mathcal{C}=\left\{\left(U_{1}, R_{1}\right),\left(U_{2}, R_{2}\right), \ldots,\left(U_{T}, R_{T}\right)\right\}$ denote a complete dialogue with $T$ turns, where $U$ and $R$ are the messages from the user and the system.
$\boldsymbol{d}=\left\{d_{1}, \ldots, d_{M}\right\}$ represents the $M$ domains supported by the dialog system, and each domain $d_{m}$ corresponds to a database $\mathcal{DB}_{d_m}$ containing many entries, each of which consists of multiple attributes.
Moreover, databases in many TOD systems predefine a set of dedicated API interfaces (i.e., intents) for querying or updating, whose calling parameters consist of specific attribute-value pairs.
In this work, we refer to domain-attribute tuples as slots, typically categorized into \textit{informable slots} that users need to specify and the system needs to track, and \textit{requestable slots} about which users can make inquiries.
Given a set of all \textit{informable slots} $\mathcal{S}=\left\{S_{1}, S_{2}, \ldots, S_{J}\right\}$, the dialog state at turn $t$ is defined as $B_{t}=\left\{S_{j}: v_{j}^{(t)} \mid 1 \leq j \leq J, v_{j}^{(t)} \in \mathcal{V}_{j}\right\}$, where $\mathcal{V}_{j}$ is the value space of slot $S_{j}$.

By putting the value spaces of all slots together, we construct the global state space $\mathbb{O}=\left\{\left(S_{1}, \mathcal{V}_{1}\right),\left(S_{2}, \mathcal{V}_{2}\right), \ldots,\left(S_{J}, \mathcal{V}_{J}\right)\right\}$, also known as the \textit{ontology}.
The \textit{\textbf{schema}} of a TOD system, which structurally describes the applicable scenarios and services that the system can provide, encompasses the subordinate slots and intents of each domain, the slots associated with the intent, and the \textit{ontology}.

\subsubsection{Modularized task-oriented dialog systems}\label{sec:modularized_tod}
In a modularized TOD system, each turn of interaction with the user requires understanding user utterances, tracking dialog states, querying the databases when necessary, making decisions, and replying to the user in natural language, where the execution of each task depends on the output of the previous task, known as the task flow.
At each turn $t$, the system first applies a natural language understanding (NLU) module to identify the domains,  intents, and slot-value pairs involved in $U_{t}$:
\begin{equation}
\begin{aligned}
    \left\{(d^{(t)}_{k}, i^{(t)}_{k}, s^{(t)}_{k})\right\}^{K}_{k=1} &=\text{NLU}\left(U_{t}\right),
\end{aligned}
\end{equation}
where $K$ is the number of identified intents, $i^{(t)}_{k}$ is the $k$-th intent, $d^{(t)}_{k}$ and $s^{(t)}_{k}$ are the domain and slot-value pairs corresponding to $i^{(t)}_{k}$, respectively.
Then, conditioned on the current dialog context $\mathcal{C}_{t}=\left\{\left(U_{1}, R_{1}\right), \ldots, \left(U_{t-1}, R_{t-1}\right), U_{t}\right\}$ and the previous dialog state $B_{t-1}$, a dialog state tracking (DST) module maintains the dialog state $B_{t}$ by modifying existing slots, adding new slot-value pairs, or deleting invalid slots in $B_{t-1}$:
\begin{equation}
\begin{aligned}
    B_{t} &=\text {DST}\left(\mathcal{C}_{t}, B_{t-1}\right).
\end{aligned}
\end{equation}
The predicted $B_{t}$ is then used to interact with the database when necessary.
Concretely, for each intent $i^{(t)}_{k}$, the dialog system extracts the slot-value pairs required for the API call from $B_{t}$ and then queries the corresponding database $\mathcal{DB}_{d^{(t)}_{k}}$ to obtain the results $E^{(t)}_{d^{(t)}_{k}}$.
Subsequently, a dialog policy (POL) module plans the next action based on the DB results and dialog state, such as recommending entities that meet user needs, requesting more information from the user when conditions are insufficient, and informing the user of the relevant information requested:
\begin{equation}
\begin{aligned}
    A_{t} &=\text{POL}\left(\mathcal{C}_{t}, B_{t}, E^{(t)}\right),
\end{aligned}
\end{equation}
where $A_{t}$ is a set of $(domain, act, slot)$ triplets, and $act$ indicates the operation applied to the slot, such as informing the slot value.
Finally, a natural language generation (NLG) module transforms the action $A_{t}$ into a fluent and informative response $R_{t}$:
\begin{equation}
\begin{aligned}
    R_{t} &=\text{NLG}\left(\mathcal{C}_{t}, A_{t}\right).
\end{aligned}
\end{equation}
For the training of a modularized TOD system, each module is individually designed and optimized on its respective task data.

\subsubsection{End-to-end task-oriented dialog modeling}
Separately modeling each module of a TOD system can lead to error propagation, where errors made by previous modules are continuously accumulated and amplified in subsequent modules, causing the system's behavior to deviate significantly from the expected.
To alleviate the error propagation problem, much work proposes various joint training mechanisms on multiple modules~\citep{DBLP:conf/acl/ChenCQYW19,DBLP:conf/acl/WangTWQY20,DBLP:conf/aaai/ZhangOY20}, enabling a single model or multiple models to learn to solve multiple tasks together.
For instance, \citet{DBLP:conf/aaai/LiangTCY20} jointly train four decoders to sequentially perform NLU, DST, POL, and NLG tasks, while \citet{DBLP:conf/emnlp/Lee21} optimizes an auto-regressive decoder to generate system action and response.

Recently, PLMs such as GPT-2~\citep{radford2019language}, T5~\citep{DBLP:journals/jmlr/RaffelSRLNMZLL20}, and UniLM~\citep{DBLP:conf/nips/00040WWLWGZH19} have been introduced into TOD systems to simplify system design and improve performance.
PPTOD~\citep{DBLP:conf/acl/SuSMG0LZ22} formalizes all TOD subtasks into the text-to-text format and trains a unified T5 model to generate the corresponding task outputs based on different task prefixes.
On the other hand, PLMs also have pushed joint modeling towards a more end-to-end manner, allowing a single model to efficiently execute the entire task flow of a TOD system for each dialog turn.
\citet{DBLP:conf/nips/Hosseini-AslMWY20} and \citet{DBLP:journals/tacl/PengLLSLG21} sequentially concatenate various TOD subtasks and database results in the causal decoding mode of auto-regressive models, allowing a single model to complete multiple tasks and more effectively extract the relationships between these tasks.
Formally, the learning objective of end-to-end TOD joint modeling for each turn $t$ is to minimize the negative log-likelihood of task outputs given the dialog context $\mathcal{C}_{t}$:
\begin{equation}
\begin{aligned}
    \mathcal{L}_{\text{turn}} &= -\log p_{\theta}\left(R_{t}, A_{t}, E^{(t)}, B_{t} \mid \mathcal{C}_{t}\right) \\
    &= -\log p_{\theta}\left(B_{t} \mid \mathcal{C}_{t}\right) \cdot p_{\theta}\left(A_{t} \mid \mathcal{C}_{t}, B_{t}, E^{(t)}\right) \cdot p_{\theta}\left(R_{t} \mid \mathcal{C}_{t}, B_{t}, E^{(t)}, A_{t}\right),
\end{aligned}
\end{equation}
where $\theta$ is the model parameters to be optimized and $E^{(t)}$ is derived from the database API calls as described in Section~\ref{sec:modularized_tod}.
UBAR~\citep{DBLP:conf/aaai/YangLQ21} further extends the end-to-end modeling to a more practical TOD setting by modeling on the dialog session level, where the intermediate task annotations of previous turns are also included in the dialog history during training.
Thus, the loss function of session-level end-to-end TOD modeling can be defined as:
\begin{equation}
\begin{aligned}
    \mathcal{L}_{\text{session}} &= - \sum_{t=1}^{T} \log p_{\theta}\left(R_{t}, A_{t}, E^{(t)}, B_{t} \mid \mathcal{H}_{t}\right), \\
    \mathcal{H}_{t} &= \begin{cases}
        \mathcal{H}_{t-1} \oplus \left\{B_{t-1}, E^{(t-1)}, A_{t-1}, R_{t-1}, U_{t}\right\}, & \text{if } t > 1; \\
        \left\{U_{1}\right\}, & \text{if } t = 1,
    \end{cases}
\end{aligned}
\end{equation}
% 表示字符串拼接操作
where the operator $\oplus$ represents the string concatenation operation.
This allows the model to mine more characteristics for each task from the coherent session-level workflow, such as incremental updates of dialog states, the coherence of action decisions, and the consistency of response generation.

\subsection{Task-oriented dialog datasets}
Benefiting from the broad prospects of intelligent assistants, task-oriented dialog systems have received widespread attention in recent years, and many task-oriented dialog datasets have been proposed to verify the system designs~\citep{DBLP:conf/acl/MrksicSWTY17,DBLP:conf/eacl/Rojas-BarahonaG17,DBLP:conf/emnlp/BudzianowskiWTC18,DBLP:journals/corr/abs-2009-13570,DBLP:conf/emnlp/ByrneKSNGDYDKC19,DBLP:conf/aaai/RastogiZSGK20,DBLP:conf/nips/LinMW0JHSF21}.
These datasets can be roughly divided into three categories: (1) language understanding datasets, (2) dialogues with modular annotations, and (3) dialogues with end-to-end annotations.
Language understanding datasets label the intents or slots in user utterances, aiding in the development of the natural language understanding module.
A complete task-oriented dialogue system typically needs to sequentially accomplish user utterance understanding, dialogue state tracking, database querying, system action decision-making, and response generation.
Therefore, multi-turn dialog datasets with intermediate task annotations have been introduced, where each turn of the dialog contains not only user utterances and system responses but also dialog states, database query results, etc.
End-to-end annotated dialog datasets, on the other hand, only provide the underlying databases for each dialog, requiring the dialog system to directly retrieve the databases to reply to the user.

In this work, we focus on modeling modularized task-oriented dialogue systems in a task-flow manner.
We carefully select \NUMDATASET{} task-oriented dialog datasets to construct the instruction-tuning corpus, including four language understanding datasets and seven multi-turn dialog datasets with modular annotations.
Table~\ref{table:datasets} shows the statistics of each dataset and the involved dialog tasks.

\subsection{Backbone models}
\label{sec:backbone_and_ift}
\bmhead{Backbone models} Both \llama{} series models open-sourced by Meta AI\footnote{\url{https://ai.meta.com/meta-ai/}} and Qwen2.5 series developed by Alibaba Cloud\footnote{\url{https://qwen.ai/}} are built on the decoder-only transformer architecture.
% The \llama{} series models, open-sourced by Meta AI\footnote{\url{https://ai.meta.com/meta-ai/}}, are built on the decoder-only transformer architecture.
These models are pre-trained on a large-scale text corpus sourced from publicly available data using a straightforward auto-regressive language modeling objective.
The \llama{} and Qwen2.5 models have demonstrated a powerful ability to comprehend context and a great potential in faithfully following TOD-related task instructions.

In this work, we use \llamaSmall{} as the backbone model, which comprises 32 transformer blocks, each with 32 attention heads and a hidden dimension of 4096.
\hlyellow{Furthermore, to validate the generalizability of our proposed framework, we introduce three additional models from the Qwen2.5-Instruct series~\citep{DBLP:journals/corr/abs-2412-15115} with varying parameter counts: 0.5B, 1.5B, and 3B.
This allows us to assess the effectiveness and scalability of the framework across different model sizes and architectures, thereby providing a more comprehensive evaluation of its performance.
The pre-training configurations of \llama{} 2 and Qwen2.5 enable the models to handle input sequences of up to 4096 and 32K tokens, respectively.}

\bmhead{Instruction fine-tuning} \hlyellow{Instruction fine-tuning (IFT)~\citep{DBLP:conf/iclr/WeiBZGYLDDL22} is a supervised learning paradigm employed to enhance the capabilities of pre-trained language models (PLMs). 
Following the initial large-scale, self-supervised pre-training phase, instruction fine-tuning further trains the model on a curated dataset composed of (instruction, input, output) triples.
These instructions are typically formulated in natural language and describe a wide variety of tasks (e.g., summarization, question answering, translation, reasoning).

The primary objective of this process is to teach the model to generalize across different tasks by understanding and following human-provided commands.
By being exposed to a diverse set of explicit instructions and their corresponding high-quality responses, the methodology has been shown to be highly effective in aligning models with human preferences~\citep{DBLP:conf/nips/Ouyang0JAWMZASR22}.
This significantly improves its zero-shot and few-shot performance on unseen tasks~\citep{DBLP:conf/iclr/WeiBZGYLDDL22}, making it a more versatile and practical tool that can be prompted to perform new functions without requiring task-specific training.}

\section{Methodology}\label{sec:methodology}
Overall, we propose an instruction-tuning paradigm for task-oriented dialogs, upon which a generalizable TOD system \ours{} is built via session-level end-to-end modeling on dialogues with modular annotations.
To enhance the applicability of \ours{} to a diverse set of TOD tasks and a wide range of domains, we employ three key mechanisms to fully leverage multiple TOD datasets with partially overlapping tasks, including an instruction-aware mechanism, a schema-aware mechanism, and a session-level end-to-end modeling mechanism.
In the subsequent sections, we will detail (1) how to construct an instruction-tuning corpus from various TOD datasets in a unified manner, and (2) how to leverage this corpus to accomplish end-to-end TOD modeling.

% 前人工作已经揭示了，已标注的TOD数据可以加速PLMs的对话预训练，进而提升结果PCMs在各种下游TOD任务上的性能。
% As revealed by previous work, labeled TOD data can be used to accelerate the dialog pre-training of PLMs, and then improve the performance of the resulting PCMs on various downstream TOD tasks~\citep{DBLP:conf/acl/SuSMG0LZ22,DBLP:conf/sigir/HeDYSHSL22}.

% \subsection{Generative task-oriented dialog tasks}
% 定制为生成式对话任务

\begin{figure*}[t]
    \centering
    \includegraphics[width=1.0\textwidth]{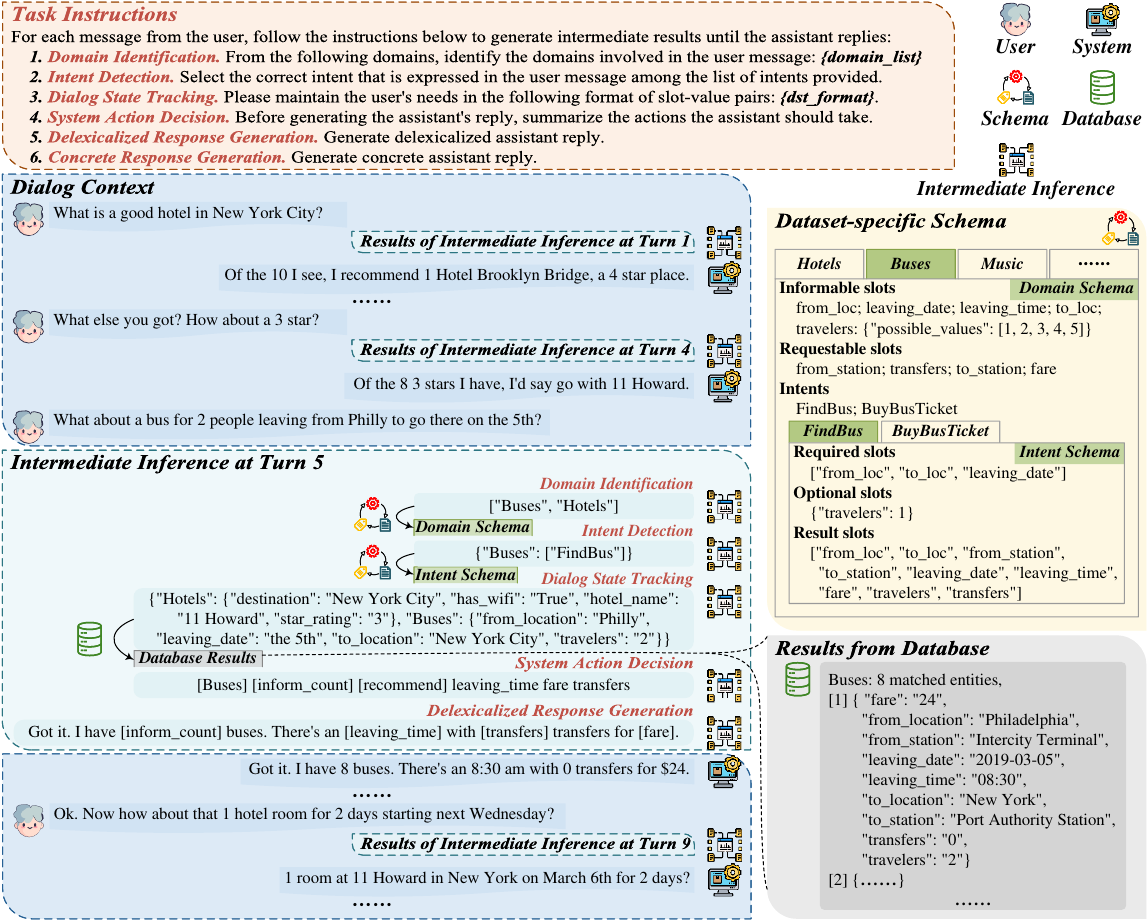}
    \caption{Illustration of the instruction-tuning paradigm for task-oriented dialog systems}
    \label{fig:overview}
\end{figure*}

\subsection{Alignment between dialog scenarios and annotations}\label{sec:instruction_schema_alignment}
Fig.~\ref{fig:overview} provides an illustrative example of transforming a dialogue session from the SGD dataset into instruction-tuning data, showcasing the comprehensive process of session-level end-to-end modeling and inference.

\begin{table}[t]
\begin{center}
\begin{minipage}{\textwidth}
    \setlength{\tabcolsep}{1.0mm}
    \caption{\hlyellow{Instruction template variations across datasets. The \textbf{Full Task Instructions} part presents example output for each task instruction. Note that the example output is not included in the instruction template.}}\label{table:inst_variation}
    \begin{tabular*}{\linewidth}{@{\extracolsep{\fill}}lp{0.85\linewidth}@{\extracolsep{\fill}}}
        \toprule
        \multicolumn{2}{@{}l@{}}{\textbf{System Instruction}} \\
        \multicolumn{2}{@{}l@{}}{\footnotesize\makecell[l]{\setTextBG{green!10}{~Please act as an AI assistant to interact with the user in a task-oriented dialogue scenario to} \\\setTextBG{green!10}{~meet his/her needs.}\\\setTextBG{green!10}{~For each message from the user, follow the instructions below to generate intermediate results} \\\setTextBG{green!10}{~until the assistant replies:}}} \\
        \cdashline{1-2}[1.5pt/3pt] \\ [-2ex]
        \multicolumn{2}{@{}l@{}}{\textbf{Full Task Instructions}} \\
        \footnotesize\textbf{~DI} & \footnotesize \setTextBG{green!10}{Please identify the domains involved in the user message from: \texttt{\textit{\{domain\_list\}}}.} \\
        \multicolumn{2}{@{}l@{}}{\footnotesize ~~\texttt{\textit{Example}: }\texttt{\textit{["banking"]}; \textit{["hotel", "train"]}}} \\
        \footnotesize\textbf{~ID} & \footnotesize \setTextBG{green!10}{Select the correct intent(s) expressed in the user text among the provided intents.} \\
        \multicolumn{2}{@{}l@{}}{\footnotesize ~~\texttt{\textit{Example}: }\texttt{\textit{\{"RentalCars": ["ReserveARentalCar"], "Media": ["MovieSearchGenre"]\}}}} \\
        \footnotesize\textbf{~DST} & \footnotesize \setTextBG{green!10}{Please maintain the user's needs from the beginning of the dialogue to the present in the following format of slot-value pairs: \texttt{\textit{\{dst\_format\}}}.} \\
        \multicolumn{2}{@{}l@{}}{\footnotesize ~~\texttt{\textit{Example}: }\texttt{\textit{\{"hotel": \{"pricerange": "cheap", "area": "west"\}, "train": \{"dest": "norwich"\}\}}}} \\
        \footnotesize\textbf{~SAD} & \footnotesize \setTextBG{green!10}{Before generating the assistant's reply, summarize the system action decisions.} \\
        \multicolumn{2}{@{}l@{}}{\footnotesize ~~\texttt{\textit{Example}: }\texttt{\textit{[hotel] [recommend] name [inform] type stars parking internet}}} \\
        \footnotesize\textbf{~Delex RG} & \footnotesize \setTextBG{green!10}{Generate delexicalized assistant reply.} \\
        \multicolumn{2}{@{}l@{}}{\footnotesize ~~\texttt{\textit{Example}: }\texttt{\textit{How about [value\_name]? It has free parking, and it's a [value\_stars]-star hotel.}}} \\
        \footnotesize\textbf{~Conc RG} & \footnotesize \setTextBG{green!10}{Generate concrete assistant reply.} \\
        \multicolumn{2}{@{}l@{}}{\footnotesize ~~\texttt{\textit{Example}: }\texttt{\textit{How about University Arms Hotel? It has free parking, and it's a 4-star hotel.}}} \\
        \midrule
        \textbf{Dataset} & \textbf{System Prompt} \\
        \midrule
        CamRest & \texttt{System Instruction} $+$ \ding{172} DI \ding{173} DST \ding{174} Delex RG \\
        In-Car & \texttt{System Instruction} $+$ \ding{172} DI \ding{173} DST \ding{174} Delex RG \ding{175} Conc RG \\
        MultiWOZ & \texttt{System Instruction} $+$ \ding{172} DI \ding{173} DST \ding{174} SAD \ding{175} Delex RG \ding{176} Conc RG \\
        SGD & \texttt{System Instruction} $+$ \ding{172} DI \ding{173} ID \ding{174} DST \ding{175} SAD \ding{176} Delex RG \ding{177} Conc RG \\
        Frames & \texttt{System Instruction} $+$ \ding{172} DI \ding{173} DST \ding{174} Conc RG \\
        \cdashline{1-2}[1.5pt/3pt] \\ [-2ex]
        \multirow{2}{*}{BiToD} & \texttt{System Instruction} $+$ \ding{172} DI \ding{173} DST \ding{174} SAD \ding{175} Delex RG \ding{176} Conc RG \\
        & \footnotesize\makecell[l]{\texttt{\textit{dst\_format}: \{} \\ \quad\texttt{"relations": ["equal\_to", "at\_least", "not", "one\_of"],} \\ \quad\texttt{"examples": \{"slot1": "one\_of(val1, val2)", "slot2": "equal\_to(val3)"\}\}}} \\
        \cdashline{1-2}[1.5pt/3pt] \\ [-2ex]
        % \texttt{dst\_format: \{"format": \{"relations": ["equal\_to", "at\_least", "not", "one\_of"], "\{slot\_name\}": "\{relation\}(value(s))"\}, "examples": \{"slot\_1": "one\_of(value\_11 , value\_12)", "slot\_2": "equal\_to(value\_21)"\}\}}
        STAR & \texttt{System Instruction} $+$ \ding{172} DI \ding{173} ID \ding{174} SAD \ding{175} Conc RG \\
        BANKING & \texttt{System Instruction} $+$ \ding{172} ID \\
        CLINC & \texttt{System Instruction} $+$ \ding{172} DI \ding{173} ID \\
        HWU & \texttt{System Instruction} $+$ \ding{172} DI \ding{173} ID \\
        SNIPS & \texttt{System Instruction} $+$ \ding{172} DI \ding{173} DST \\
        \bottomrule
    \end{tabular*}
\footnotetext{The numbers (\ding{172}-\ding{177}) denote the sequential order of the instructions and the corresponding tasks to be executed.}
\footnotetext{Except for BiToD, the \texttt{\textit{dst\_format}} of the datasets containing DST annotations is \texttt{\textit{\{"format": \{"\{slot\_name\}": "\{slot\_value\}"\}, "examples": \{"slot1": "val1", "slot2": "val2"\}\}}}.}
\end{minipage}
\end{center}
\end{table}

\bmhead{Instruction-aware mechanism}
PPTOD~\citep{DBLP:conf/acl/SuSMG0LZ22} jointly learns all TOD tasks by recasting them as text generation problems with task-specific prompts, without explicitly modeling the relationships between tasks.
Although existing methods~\citep{DBLP:conf/nips/Hosseini-AslMWY20,DBLP:journals/tacl/PengLLSLG21,DBLP:conf/aaai/YangLQ21} successfully achieve end-to-end modeling of a fixed task flow in TOD systems, they overlook the diversity in task sets and annotation formats that different TOD datasets may present.
Motivated by previous work, we establish a series of \textbf{\textit{task instructions}} prior to initiating a conversation.
These instructions delineate detailed guidelines for a TOD system, for example, generating dialogue states according to a specified annotation format, and the order in which the TOD system should complete each task.
\hlyellow{It is worth noting that there are differences in instruction templates across datasets, which are summarized in Table~\ref {table:inst_variation}.}
This variability allows the instruction-tuned system to adapt flexibly to diverse task flows and develop instruction awareness.

\bmhead{Schema-aware mechanism}
The schemas, defined in Section~\ref{sec:notations}, are typically incompatible across different datasets, posing a challenge for existing methods to make consistent predictions with the corresponding schema.
Different from specialized TOD systems tailored for specific datasets, recent research has shifted towards building generalizable subtask modules by explicitly providing the corresponding schema in the task input.
\citet{DBLP:conf/emnlp/ZhangPLZM23} constrain the LLM to generate dialog states according to the given slot names and possible values, and \citet{DBLP:conf/emnlp/00010O21} supervise T5 to understand slot descriptions and generate corresponding slot values.
Recognizing the critical role of schema information in general-purpose TOD systems, we integrate it into our dialogue context modeling to advance LLMs to generate outputs that are aware of the specific schema.
As illustrated in Fig.~\ref{fig:overview}, once the domains associated with each user utterance are identified, the schema of each activated domain — encompassing its slots, possible values, and list of intents — is directly appended to the domain recognition results.
In this way, subsequent tasks, such as intent detection or dialog state tracking, can utilize the dialog history to identify intents and slots mentioned by users within the predefined schema, thereby preventing the system from yielding unexpected outcomes.
Similarly, if the dataset predefines the schema for intents, the schemas corresponding to the recognized intents will also be incorporated into the intent recognition results.

\subsection{Session-level end-to-end modeling}
\label{sec:session_level_e2e_modeling}
Turn-level end-to-end TOD modeling is centered on jointly learning the task flow for each user message, taking both the current user message $U_{t}$ and the dialogue history $\mathcal{C}_{1:t}= \left\{\left(U_{1}, R_{1}\right), \ldots, \left(U_{t-1}, R_{t-1}\right)\right\}$ as input.
However, this turn-level modeling restricts the system from accessing historical execution records for each task, which could compromise the behavioral consistency vital to the functionality of a practical TOD system.
Following UBAR~\citep{DBLP:conf/aaai/YangLQ21}, we preserve the task flow execution results of each dialogue turn within the dialogue history, enabling session-level end-to-end TOD modeling.
Specifically, for the $t$-th turn of the dialogue in Fig.~\ref{fig:overview}, our instruction-tuning framework defines the dialogue context as:
\begin{equation}
\begin{aligned}\label{eq:dialogue_context}
    \mathcal{H}_{t} &= \begin{cases}
        \mathcal{H}_{t-1} \oplus \left\{d^{(t)}, i^{(t)}, B_{t-1}, E^{(t-1)}, A_{t-1}, R_{t-1}, U_{t}\right\}, & \text{if } t > 1; \\
        \left\{\boldsymbol{inst}, U_{1}\right\}, & \text{if } t = 1.
    \end{cases}
\end{aligned}
\end{equation}
In Equation~\ref{eq:dialogue_context}, $\boldsymbol{inst}$ denotes the task instructions, while $d^{(t)}$ and $i^{(t)}$ respectively represent the domain and intent recognition results of the user message $U_{t}$, with the corresponding schemas included in both.
Moving forward, we will describe the implementation details of each TOD-related task and formalize the process of session-level end-to-end TOD modeling process.

\bmhead{Task: Domain identification}
Given the impracticality of incorporating the schemas for all domains supported by a TOD system within the context, it is essential to first identify the domains relevant to the user utterance at each turn.
We approach the domain identification task — a multi-label classification challenge — as a generative problem, where domain labels are directly produced as a serialized Python list to facilitate easy parsing.
The learning objective of generating the domain label sequence $d^{(t)}$ is defined using the standard negative log-likelihood loss:
\begin{equation}
\begin{aligned}\label{eq:domain_loss}
    \mathcal{L}_{\text{domain}} &= -\log p_{\theta}\left(d^{(t)} \mid \mathcal{H}_{t}\right) = -\sum_{l=1}^{L_{d}} \log p_{\theta}\left(d^{(t)}_{l} \mid \mathcal{H}_{t}, d^{(t)}_{<l}\right),
\end{aligned}
\end{equation}
where $L_{d}$ is the number of tokens derived from tokenizing the sequence $d^{(t)}$, and $d^{(t)}_{<l}$ denotes all tokens that precede the $l$-th token in $d^{(t)}$.
Subsequently, the active domains can be easily parsed from the generated $d^{(t)}$ and the corresponding schemas are appended to $d^{(t)}$:
\begin{equation}
\begin{aligned}
    \tilde{d}^{(t)} &= \text{parse}\left(d^{(t)}\right), \\
    d^{(t)} &= d^{(t)} \oplus \left\{\text{schema}_{dom}\right\}_{dom \in \tilde{d}^{(t)}}.
\end{aligned}
\end{equation}
Here, $\tilde{d}^{(t)}$ represents the set of identified domains obtained through the parsing function $\text{parse}(\cdot)$, such as deserialization.

\bmhead{Task: Intent detection}
To accommodate multi-intent scenarios, wherein users express several intents within a single domain or across multiple domains in a single turn, we utilize a Python dictionary structure to represent intents.
This structure is formatted as $\left\{\text{domain}_{1}: \text{intent\_list}_{1}, \text{domain}_{2}: \text{intent\_list}_{2}, \ldots\right\}$.
The training objective of intent detection is computed as:
\begin{equation}
\begin{aligned}\label{eq:intent_loss}
    \mathcal{L}_{\text{intent}} &= -\log p_{\theta}\left(i^{(t)} \mid \mathcal{H}_{t}, d^{(t)}\right) = -\sum_{l=1}^{L_{i}} \log p_{\theta}\left(i^{(t)}_{l} \mid \mathcal{H}_{t}, d^{(t)}, i^{(t)}_{<l}\right),
\end{aligned}
\end{equation}
where $i^{(t)}$ is the serialized intent label sequence of length $L_{i}$.
Similar to domain identification, the schemas of the parsed intents $\tilde{i}^{(t)}$ are appended to $i^{(t)}$.

\begin{algorithm}[t]
    \caption{Schema management}\label{alg:schema_management}
    \begin{algorithmic}[1]
        \Require Dialog session $\mathcal{C}$ with $T$ turns
        \Require Schema type $type \in \{\text{domain}, \text{intent}\}$ and schema set $\mathbb{S}_{type}$
        \Require Maximum number of history turns $W$
        \State Initialize an empty mapping $\mathcal{M}_{type} \Leftarrow \varnothing$
        \For{each turn $\left(U_{t}, R_{t}\right)$ in $\mathcal{C}$}
            \State Obtain parsed labels $y^{(t)}$ of $U_{t}$, such as domains $\tilde{d}^{(t)}$ and intents $\tilde{i}^{(t)}$
            \For{each label $y_{k}^{(t)}$ in $y^{(t)}$}
                \If{$y_{k}^{(t)} \notin \mathcal{M}_{type}$ or $\mathcal{M}_{type}\left[y_{k}^{(t)}\right] \gt W$}
                    \State Append schema $\mathbb{S}_{type}\left[y_{k}^{(t)}\right]$ to corresponding recognition results
                    \State $\mathcal{M}_{type}\left[y_{k}^{(t)}\right] \Leftarrow 0$
                \EndIf
            \EndFor
            % \For{each intent $\tilde{i}^{(t)}_{k}$ in $\tilde{i}^{(t)}$}
            %     \If{$\tilde{i}^{(t)}_{k} \notin \mathcal{M}_{int}$ or $\mathcal{M}_{int}\left[\tilde{i}^{(t)}_{k}\right] \gt W$}
            %         \State Append schema $\mathbb{S}_{int}\left[\tilde{i}^{(t)}_{k}\right]$ to intent recognition results
            %         \State $\mathcal{M}_{int}\left[\tilde{i}^{(t)}_{k}\right] \Leftarrow 0$
            %     \EndIf
            % \EndFor
            \For{$e$ in $\mathcal{M}_{type}$}
                \State $\mathcal{M}_{type}\left[e\right] \Leftarrow \mathcal{M}_{type}\left[e\right] + 1$
            \EndFor
            % \For{$i$ in $\mathcal{M}_{int}$}
            %     \State $\mathcal{M}_{int}\left[i\right] \Leftarrow \mathcal{M}_{int}\left[i\right] + 1$
            % \EndFor
        \EndFor
    \end{algorithmic}
\end{algorithm}

\bmhead{Schema management}
In practice, when multiple dialogue turns discuss the same topic, the same schema may be repeated numerous times within the dialogue context.
Therefore, we employ a schema management algorithm to control the incorporation of schemas during data construction and end-to-end inference.
This technique allows schemas to be shared across multiple dialogue turns, as detailed in Algorithm~\ref{alg:schema_management}.

\bmhead{Task: Dialog state tracking}
As illustrated in Fig.~\ref{fig:overview}, the dialog state is represented as a serialized nested dictionary, with the state of each domain comprising slot-value pairs.
The generation loss of the dialog state $B_{t}$:
\begin{equation}
\begin{aligned}\label{eq:dst_loss}
    \mathcal{L}_{\text{dst}} &= -\sum_{l=1}^{L_{B}} \log p_{\theta}\left(B_{t, l} \mid \mathcal{H}_{t}, d^{(t)}, i^{(t)}, B_{t, <l}\right).
\end{aligned}
\end{equation}
The parsed dialog state is then used to interact with the underlying databases.
Unlike prior work, our database query results $E^{(t)}$ not only include the number of matched entities or the success of the API call, but also contain detailed API outcomes.

\bmhead{Task: System action decision and response generation}
Lastly, the system action $A_{t}$ and response $R_{t}$ are sequentially produced, grounded in the dialogue context $\mathcal{H}_{t}$ and the outputs of previous tasks.
The corresponding loss functions are:
\begin{align}
    \mathcal{L}_{\text{pol}} &= -\sum_{l=1}^{L_{A}} \log p_{\theta}\left(A_{t, l} \mid \mathcal{H}_{t}, d^{(t)}, i^{(t)}, B_{t}, E^{(t)}, A_{t, <l}\right),\label{eq:pol_loss} \\
    \mathcal{L}_{\text{nlg}} &= -\sum_{l=1}^{L_{R}} \log p_{\theta}\left(R_{t, l} \mid \mathcal{H}_{t}, d^{(t)}, i^{(t)}, B_{t}, E^{(t)}, A_{t}, R_{t, <l}\right).\label{eq:nlg_loss}
\end{align}

\subsection{Instruction-tuning and inference}\label{sec:e2e_ft_and_eval}
In summary, the core of our proposed instruction-tuning architecture lies in three key mechanisms: (1) the instruction-aware mechanism, which unifies TOD datasets with varying task sets, ensuring the fine-tuned model can follow instructions to accomplish the corresponding tasks, (2) the schema-aware mechanism, which aligns the dialog-specific schema (e.g., intent names, slot names) with the task annotations of each dialog, enabling the model to make predictions consistent with the given schema, and (3) the session-level end-to-end modeling mechanism, which captures the information flow between tasks and the behavioral changes of each task across dialog turns.

\bmhead{Full learning objective}
Due to the considerable overhead of annotating task-oriented dialog data, not all datasets are as comprehensively annotated as the SGD dataset, which includes all five TOD tasks $\mathcal{T} = \left\{\text{domain}, \text{intent}, \text{dst}, \text{pol}, \text{nlg}\right\}$.
This results in heterogeneous dialog corpora.
In this scenario, we aim to employ instruction-tuning across various heterogeneous datasets to develop a generalizable TOD system that adheres closely to task instructions and schemas.
Let $\mathcal{D} = \union_{i=1}^{N} \mathcal{D}_{i}$ denote the instruction-tuning corpus, which comprises $N$ TOD datasets.
Each dataset $\mathcal{D}_{i}$ includes a subset $\mathcal{T}_{i}$ of total tasks $\mathcal{T}$ and is constructed according to our instruction-tuning paradigm.
The total loss $\mathcal{L}_{\theta}\left(\mathcal{D}\right)$ over our instruction-tuning corpus $\mathcal{D}$ combines multiple training objectives from~\eqref{eq:domain_loss} to~\eqref{eq:nlg_loss}:
\begin{equation}
\begin{aligned}
    \mathcal{L}_{\theta}\left(\mathcal{D}\right) &= \sum_{i=1}^{N} \sum_{n=1}^{\lvert \mathcal{D}_{i} \rvert} \sum_{task \in \mathcal{T}_{i}} \mathcal{L}_{task}\left(x_{n}\right),
\end{aligned}
\end{equation}
where $x_{n}$ is the $n$-th dialog in the $i$-th dataset.
In the instruction-tuning stage, the model parameters $\theta$ are learned via mini-batch optimization, as detailed in Section~\ref{sec:implementation_details}.

\bmhead{Inference}
For a new dialogue from dataset $\mathcal{D}_{i}$, the system is first fed with the task instructions $\boldsymbol{inst}$ defined by the task subset $\mathcal{T}_{i}$ and the user utterance $U_{1}$ to initiate the inference process.
Then the generated content of each task is retained within the task flow of the current turn, and the task flow execution results of each turn are preserved in the dialogue context to facilitate session-level end-to-end TOD inference, as illustrated in Fig.~\ref{fig:overview}.

\bmhead{Error Handling}
\hlyellow{To ensure system robustness against malformed or unparsable outputs from the language understanding module, particularly in complex multi-intent scenarios, we employ a strict error-handling protocol during the parsing stage.
Our framework utilizes a standard JSON parser to convert the model's textual output into structured data formats, such as nested dictionaries and lists, which are essential for schema retrieval and database interaction.

In instances where the output is syntactically incorrect and fails to parse, the system is designed to return a default empty structure (e.g., an empty dictionary or list) instead of raising an exception.
This prevents system crashes and allows the dialogue to proceed.
Crucially, the original, unparsed string generated by the model is always retained in the dialogue context.
This design choice ensures no information is lost and provides an opportunity for downstream error analysis or recovery mechanisms.
Empirically, we have found that such parsing failures are infrequent, even with smaller-scale models.}

\bmhead{Context Management} \hlyellow{As stated in Section~\ref{sec:session_level_e2e_modeling}, our model \textsc{ESAinsTOD} is fine-tuned and evaluated in a session-level end-to-end manner.
In this paradigm, the context ideally includes all previous user messages, system outputs for various dialogue tasks, and potentially extensive schema definitions and database results.
For multi-turn dialogues, this concatenated input can certainly exceed the model's token limit.
To manage inputs that exceed the model's maximum token limit (e.g., 4096 tokens for LLaMA 2 7B), we employ a straightforward yet effective sliding-window strategy for truncation as follows:}
\begin{itemize}
    \item \hlyellow{\textbf{During Fine-tuning:} For long dialogue sessions, we segment them into multiple training samples using a sliding window. The window is carefully selected to ensure that any turn that would otherwise be truncated still retains sufficient historical context to be a meaningful training example. In our implementation, we allocate up to 60\% of the maximum token length for the preceding dialogue history.}
    \item \hlyellow{\textbf{During Inference:} At each turn of the conversation, we preserve the most recent historical turns that fit within approximately three-quarters (75\%) of the maximum token limit. The objective of retaining this sufficient context is to ensure that the model can fully model the relationship between the dialogue history and the current dialogue turn, which is crucial for coherent and context-aware responses.}
\end{itemize}
\hlyellow{Beyond the primary sliding-window approach, we have implemented two other mechanisms to further mitigate token consumption:}
\begin{itemize}
    \item \hlyellow{\textbf{Schema Management:} As presented in Algorithm~\ref{alg:schema_management}, we employ a schema management technique to reduce the token usage associated with schema definitions.}
    \item \hlyellow{\textbf{Output Deduplication:} During inference, we also reduce the token count of task results by identifying and removing repetitive sequences from overly long generated text.}
\end{itemize}
\hlyellow{In Section~\ref{sec:dialog_context}, we will analyze the impact of sliding window size on end-to-end dialogue modeling performance of \textsc{ESAinsTOD} by limiting the number of history turns.}

\section{Experiment setup}\label{sec:experiment_setup}
In this section, we will first introduce the benchmarks used in our experiments, along with the evaluation metrics and baselines for each task.
Then, we will briefly describe the implementation details of our instruction-tuning framework and our experimental settings.

\subsection{Benchmarks}
We test \ours{} on three types of task-oriented benchmarks, including language understanding, dialogue state tracking, and end-to-end dialog modeling.

\subsubsection{Language understanding}
Natural language understanding (NLU) is a fundamental task in task-oriented dialog systems, which assists in subsequent modules (i.e., dialogue management and response generation) by classifying user intents and recognizing the related slots in user utterances~\citep{tur2011spoken,DBLP:journals/pieee/YoungGTW13,DBLP:conf/ijcai/QinXC021,DBLP:journals/corr/abs-2402-03900}.

%\bmhead{Intent Recognition (IC)}
\bmhead{Evaluated datasets}
Intent detection (ID) aims to recognize the intentions expressed in the current user utterance.
We choose three popular single-turn benchmarks: \textit{BANKING77}~\citep{casanueva-etal-2020-efficient} with 77 intents in the banking domain, \textit{HWU64}~\citep{DBLP:conf/iwsds/LiuESR19} with 64 intents spanning 21 domains, and \textit{CLINC150}~\citep{DBLP:conf/emnlp/LarsonMPCLHKLLT19} with 150 intents spanning 10 domains.
% We also consider a more challenging setting, where each user utterance may not involve any intent or involve one or more intents.
% The intent task in Schema-Guided Dialogue (\textit{SGD}) Dataset~\citep{DBLP:conf/aaai/RastogiZSGK20} is adopted as our benchmark.

% 槽位填充基于预定义的槽位集合，识别用户话语中提到的槽位值。
Slot filling (SF) is to recognize the slot values mentioned in the user utterances based on a predefined set of slots.
%槽位填充数据集通常也包含意图标注，两个任务紧密耦合，例如识别用户意图有助于确定应该填充哪些槽位。
The SF datasets usually contain intent annotations as well, and the two tasks are closely coupled, e.g., recognizing user intents may help determine which slots should be filled.
Therefore, considering the interaction between the two tasks can achieve a more accurate understanding of user utterances.
\textit{SNIPS}~\citep{DBLP:journals/corr/abs-1805-10190} is a widely used benchmark, which contains 7 intent types and 72 slot labels in the home automation domain.

\bmhead{Baselines}
We compare \ours{} with several classification-based approaches, including \textbf{ConvBERT}~\citep{DBLP:journals/corr/abs-2009-13570}, \textbf{USE+ConveRT}~\citep{casanueva-etal-2020-efficient}, \textbf{Example+Observer}~\citep{DBLP:conf/naacl/MehriE21}, and \textbf{ConvFit}~\citep{DBLP:conf/emnlp/VulicSCGBCMW21}.
Among them, \textbf{Example+Observer} and \textbf{ConvFit} are similarity-based classification methods, which treat all sentences belonging to an intent as diverse surface instances, and predict the intent by finding the most similar instance to the test sentence from the training set.
On SNIPS, three joint models are compared, including \textbf{BERT-Joint}~\citep{DBLP:journals/corr/abs-1907-02884}, \textbf{Stack-Propgation+BERT}~\citep{DBLP:conf/emnlp/QinCLWL19}, and \textbf{Co-Interactive transformer+BERT}~\citep{DBLP:conf/icassp/QinLCKZ021}.

It is worth noting that most methods model intent recognition as a sentence classification task and slot filling as a sequence labeling task.
In contrast, our method directly generates the intent text and textual slot-value pairs based on user utterances and schema information, making it more flexible to adapt to new domains and settings, such as multi-intent recognition.

\bmhead{Evaluation metrics}
We adopted turn accuracy as the evaluation metric for intent recognition, which is the percentage of user turns where the model correctly predicts the active intent.
Following \citet{DBLP:conf/naacl/GooGHHCHC18} and \citet{DBLP:conf/emnlp/QinCLWL19}, SNIPS is evaluated by slot F1 score, intent accuracy, and overall accuracy.

\subsubsection{Dialogue State Tracking (DST)}
As a core component of task-oriented dialog systems, DST aims to extract and maintain user constraints consisting of slot-value pairs.
DST extends the single-turn slot filling task into the interactive process between users and dialogue agents, necessitating the model to accurately track user needs throughout the dialogue history, which encompasses various linguistic phenomena such as reference and ellipsis.

\bmhead{Evaluated datasets}
% \textit{MultiWOZ}~\citep{DBLP:conf/emnlp/BudzianowskiWTC18} and \textit{SGD}~\citep{DBLP:conf/aaai/RastogiZSGK20} are two challenging datasets for multi-domain DST, which contain many multi-domain dialogues, i.e., a dialogue session may involve domain or topic transitions multiple times.
\textit{MultiWOZ}~\citep{DBLP:conf/emnlp/BudzianowskiWTC18} is a challenging dataset for multi-domain DST, which contains many multi-domain dialogues, i.e., a dialogue session may involve domain or topic transitions multiple times.

\bmhead{Baselines}
All baselines for DST can be divided into three categories: 
1) \textit{classification-based} methods which select the value for each slot from a pre-defined candidate set, such as \textbf{TripPy}~\citep{DBLP:conf/sigdial/HeckNLGLMG20} and \textbf{FPDSC}~\citep{DBLP:conf/sigdial/ZhouWLLZ21}; 
2) \textit{generation-based} methods which directly generate slot values without being restricted to a fixed ontology, such as \textbf{SOM-DST}~\citep{DBLP:conf/acl/KimYKL20}, \textbf{SimpleTOD}~\citep{DBLP:conf/nips/Hosseini-AslMWY20}, \textbf{Seq2seq-DU}~\citep{DBLP:conf/acl/FengWL20}, \textbf{UBAR}~\citep{DBLP:conf/aaai/YangLQ21}, and \textbf{SOLOIST}~\citep{DBLP:journals/tacl/PengLLSLG21};
3) \textit{hybrid} methods which extract slot values from the dialogue history for non-categorical slots and select correct options for categorical slots, including \textbf{DSTQA}~\citep{DBLP:journals/corr/abs-1911-06192}, \textbf{DS-DST}~\citep{DBLP:conf/starsem/ZhangHWWYSX20}, and \textbf{DSS-DST}~\citep{DBLP:conf/acl/GuoSLW20}.
% For SGD,

\bmhead{Evaluation metrics}
Joint Goal Accuracy (JGA) is used by much previous work~\citep{DBLP:conf/acl/WuMHXSF19,DBLP:conf/sigdial/HeckNLGLMG20,DBLP:conf/starsem/ZhangHWWYSX20} to measure the performance of dialogue state tracking, which is the ratio of user turns that all the slot-value pairs predicted by the model exactly match the ground truth.
Different from previous methods that train and evaluate DST separately, we report the performance of DST in end-to-end modeling to verify the advantages of our instruction-tuning framework in multi-task joint modeling.

\subsubsection{End-to-end (E2E) Dialog Modeling}
E2E dialog modeling is a challenging task that can measure the overall performance of a dialog system, where an agent needs to first perform language understanding, dialogue state tracking and database (DB) query, and then decide the next system actions based on the dialogue history and DB query results, and finally generate natural language responses.
This fully end-to-end evaluation setting is closer to reality, where the system's actions and responses are not only related to the dialogue context but also conditioned on the generated dialogue state and DB query results.

\bmhead{Evaluated datasets}
Following~\citet{DBLP:conf/sigir/HeDYSHSL22}, our model is evaluated on four widely used task-oriented dialog datasets: \textit{CamRest676}~\citep{DBLP:conf/eacl/Rojas-BarahonaG17} with 676 dialogues on assisting users to find restaurants, \textit{In-Car}~\citep{DBLP:conf/sigdial/EricKCM17} where the car assistant responds to user requests about three distinct domains (i.e., calendar scheduling, weather information retrieval, and point-of-interest navigation), \textit{MultiWOZ2.0}~\citep{DBLP:conf/emnlp/BudzianowskiWTC18} and \textit{MultiWOZ2.1}~\citep{DBLP:conf/lrec/EricGPSAGKGKH20}.
\textit{MultiWOZ2.1} is a newer version of \textit{MultiWOZ2.0} with more accurate state annotations.

\bmhead{Baselines}
To demonstrate the effectiveness of our proposed \ours{} in end-to-end TOD modeling, we compare it with the following state-of-the-art baselines: 1) \textbf{SimpleTOD}, \textbf{UBAR}, \textbf{SOLOIST}, \textbf{MinTL}~\citep{DBLP:conf/emnlp/LinMWF20}, \textbf{TOP+NOD}~\citep{DBLP:journals/tacl/LiuYRB21}, and \textbf{MTTOD}~\citep{DBLP:conf/emnlp/Lee21} for MultiWOZ; 2) \textbf{SEDST}~\citep{DBLP:conf/cikm/JinLRCLZY18}, \textbf{TSCP}~\citep{DBLP:conf/acl/KanHLJRY18}, \textbf{FSDM}~\citep{DBLP:conf/sigdial/ShuMNXLZT19}, and \textbf{LABES}~\citep{DBLP:conf/emnlp/ZhangOHF20} for CamRest676 and In-Car.

\bmhead{Evaluation metrics}
The Combined Score~\citep{DBLP:conf/sigdial/MehriSE19} is a common overall metric for E2E dialog modeling, which is the weighted sum of three sub-metrics: \texttt{\textbf{Comb} = BLEU + 0.5 $\times$ (Inform Rate + Success Rate)}.
\texttt{BLEU}~\citep{DBLP:conf/acl/PapineniRWZ02} measures the similarity between the generated response and the ground truth.
\texttt{Inform Rate} and \texttt{Success Rate} reflect whether the agent offers a suitable entity and answers all the requested information in the generated response, respectively.
For \textit{CamRest676} and \textit{In-Car}, we report the \texttt{Match} and \texttt{SuccF1} from~\citet{DBLP:conf/acl/KanHLJRY18}, which correspond to \texttt{Inform Rate} and \texttt{Success Rate}, respectively.
When calculating evaluation metrics on \textit{MultiWOZ} datasets, the generated dialogue states are used to query the databases for DB results, which are then compared with the ground-truth DB results to obtain \texttt{Inform Rate}.

\subsubsection{Comparison with Pre-trained conversation models}
% \textbf{PLATO}~\citep{DBLP:conf/acl/BaoHWWW20}
Except for the above task-specific models, we also compare with three strong PCMs, including \textbf{TOD-BERT}~\citep{DBLP:conf/emnlp/WuHSX20}, \textbf{PPTOD}~\citep{DBLP:conf/acl/SuSMG0LZ22}, and \textbf{SPACE}~\citep{DBLP:conf/sigir/HeDYSHSL22}.
These PCMs are further pre-trained on large-scale TOD datasets with various well-designed pre-training objectives, and then fine-tuned on the above three tasks.
% \textbf{PLATO} with RS and response generation conditioned on the discrete latent variable
Specifically, \textbf{TOD-BERT} is pre-trained with masked language modeling (MLM) and response selection (RS), and \textbf{PPTOD} recasts all TOD-related tasks into input-output pairs on which a single multi-task model is pre-trained.
\textbf{SPACE} is a unified semi-supervised PCM learned from both labeled and unlabeled dialog corpora.
It comprises four successive components and is modeled within a TOD task flow with span-based MLM for the dialog encoder, semi-supervised contrastive learning objectives for dialog understanding and dialog policy decoders, and language modeling for the response decoder.

\subsection{Implementation details}\label{sec:implementation_details}
We utilize \llamaSmall{}~\citep{DBLP:journals/corr/abs-2307-09288} as our backbone model and perform supervised fine-turning using PyTorch~\citep{DBLP:conf/nips/PaszkeGMLBCKLGA19}, Transformers library~\citep{DBLP:conf/emnlp/WolfDSCDMCRLFDS20}, FlashAttention~\citep{DBLP:conf/nips/DaoFERR22}, and DeepSpeed~\citep{DBLP:conf/kdd/RasleyRRH20}.

\bmhead{Instruction-tuning}
\hlyellow{Given that our preliminary experiments showed the superior performance of full-parameter fine-tuning over parameter-efficient fine-tuning (specifically, LoRA~\citep{DBLP:conf/iclr/HuSWALWWC22}), we employ fully fine-tuned LLMs to build our task-oriented dialogue system.}
We set the maximum sequence length to 4096 tokens and the batch size to 64.
For multi-turn dialogues, each sample consists of multiple consecutive dialogue turns in a dialogue session, where each dialogue turn contains the user message, intermediate task results, and system response.
Long dialogues are split into multiple samples, while sufficient context is retained for each dialogue segment to ensure that the model can fully model the relationship between the dialogue history and the current dialogue turn.
For single-turn language understanding datasets, we concatenate multiple examples to fully utilize the sequence length and use a special token for separation.

\bmhead{Optimization}
During the fine-tuning phase, we use the Fused Adam optimizer with L2 regularization (commonly referred to as AdamW~\citep{DBLP:conf/iclr/LoshchilovH19}), with a weight decay set at 0.1 and gradient clipping capped at 1.0.
This optimization is applied to an autoregressive language modeling objective.
A cosine learning rate scheduler is adopted, initiating with a warmup ratio of 0.1 and reaching a peak learning rate of 5e-5.
Notably, loss calculation is restricted solely to the answer tokens for each task, ignoring all other tokens.
The fine-tuning process for Llama 2 7B on our instruction-tuning corpus is completed in approximately 6 hours over 2 epochs, using 8 NVIDIA Tesla A100 80GB GPUs.

\bmhead{Inference details}
A high-throughput Large Language Model (LLM) serving engine, named vLLM~\citep{DBLP:conf/sosp/KwonLZ0ZY0ZS23}, is employed to accelerate the inference processes in dialog systems.\footnote{\url{https://github.com/vllm-project/vllm}}
A greedy decoding strategy is used for generating task results across various benchmarks.
For each task, generation terminates either upon reaching the maximum token length or upon encountering the designated stop tokens.
To ensure the stability of the inference process, multiple inferences are performed on each sample using different random seeds, and the average metrics are reported.

\subsection{Experimental settings}
To investigate the effectiveness and generalization ability of our instruction-tuning framework, we conduct full training, low-resource, and zero-shot evaluations.
\bmhead{Full training evaluation}
Complete training sets from all datasets are mixed for fine-tuning, and the trained model is evaluated on the test sets of multiple benchmarks.
The purpose of this setting is to explore the upper performance of our framework on various task-oriented dialog benchmarks.

\bmhead{Low-resource and zero-shot evaluations}
Recent studies have paid increasing attention to the transferability of dialog models to rare or unseen dialog scenarios.
We combine the training sets from five task-oriented dialog datasets — namely, In-Car, SGD, Frames, BiTOD, and STAR — with different percentages of MultiWOZ data (i.e., 0\%, 5\%, 10\%, 20\%) for fine-tuning the backbone model, where 0\% means that the model has never seen any MultiWOZ data during training (i.e., zero-shot setting).
Subsequently, we evaluate the model on the MultiWOZ test set.
\hlyellow{The zero-shot evaluation serves to measure our framework's ability to support extending to new domains or intents without retraining from scratch.}

Furthermore, we also simulate another low-resource scenario by varying the percentage of randomly sampled data from the instruction-tuning corpus to study the effectiveness of our method in data-scarce situations.
To ensure the stability of the results in the low-resource setting, we report the average scores by sampling and training multiple times with three different random seeds.

\section{Experiment results}\label{sec:experiment_results}

In this section, we present the experimental results and analysis of our proposed method for the task-oriented dialog system.

\subsection{Main results}
The evaluation results of individual tasks under the full-data setting are shown in Tables~\ref{table:e2e_ck_full}, \ref{table:e2e_mwoz_full}, \ref{table:dst_full}, \ref{table:intent_full} and \ref{table:slu_full}.

\begin{table}[t]
\begin{center}
\begin{minipage}{\textwidth}
    \setlength{\tabcolsep}{1.6mm}
    \caption{End-to-end dialog modeling results on CamRest676 and In-Car}\label{table:e2e_ck_full}
    \begin{tabular*}{\linewidth}{@{\extracolsep{\fill}}lcccccccc@{\extracolsep{\fill}}}
        \toprule
        \makebox[0.15\textwidth][l]{\multirow{2}{*}{\textbf{Model}}} & \multicolumn{4}{@{}c@{}}{\textbf{CamRest676}} & \multicolumn{4}{@{}c@{}}{\textbf{In-Car}} \\ \cmidrule(lr){2-5} \cmidrule(lr){6-9}
            & Match & SuccF1 & BLEU & Comb & Match & SuccF1 & BLEU & Comb \\
        \midrule
        SEDST & 92.70 & 75.40 & 23.60 & 107.65 & 84.50 & 82.90 & 19.30 & 103.00 \\
        TSCP & 92.70 & 85.40 & 25.30 & 114.35 & 84.50 & 81.10 & 21.90 & 104.70 \\
        LABES & 96.40 & 83.00 & 25.50 & 115.20 & \underline{85.80} & 77.00 & 22.80 & 104.20 \\
        FSDM & 93.50 & 86.20 & \underline{25.80} & 115.65 & 84.80 & 82.10 & 21.50 & 104.95 \\
        $\text{SPACE}^\star$ & \underline{97.74} & \underline{88.24} & 23.68 & \underline{116.67} & 85.26 & \underline{83.16} & \underline{22.92} & \underline{107.13} \\
        \midrule
        {\begin{tabular}[l]{@{}c@{}}\ours{}\\[-1ex]{\tiny (\llamaSmall{})}\end{tabular}} & \textbf{98.50} & \textbf{88.45} & \textbf{26.92} & \textbf{120.39} & \textbf{90.58} & \textbf{88.09} & \textbf{27.87} & \textbf{117.21} \\
        % \makecell[l]{\ours{}\\[-1ex]{\tiny (\llama{})}} & \textbf{98.50} & \textbf{89.24} & \textbf{26.83} & \textbf{120.70} & \textbf{90.58} & \textbf{88.00} & \textbf{28.63} & \textbf{117.92} \\
        \bottomrule
    \end{tabular*}
\footnotetext{Best results are in bold, and the second best results are underlined.}
% 相关PCM被分别地在各个下游数据集上进一步微调
\footnotetext[\star]{~Corresponding pre-trained conversational model (PCM) has been further fine-tuned on each downstream dataset individually.}
\end{minipage}
\end{center}
\end{table}
    
\begin{table}[t]
\begin{center}
\begin{minipage}{\textwidth}
    \setlength{\tabcolsep}{1.0mm}
    \caption{End-to-end dialog modeling results on MultiWOZ}\label{table:e2e_mwoz_full}
    \begin{tabular*}{\linewidth}{@{\extracolsep{\fill}}lcccccccc@{\extracolsep{\fill}}}
        \toprule
        \makebox[0.15\textwidth][l]{\multirow{2}{*}{\textbf{Model}}} & \multicolumn{4}{@{}c@{}}{\textbf{MultiWOZ 2.0}} & \multicolumn{4}{@{}c@{}}{\textbf{MultiWOZ 2.1}} \\ \cmidrule(lr){2-5} \cmidrule(lr){6-9}
            & Inform & Success & BLEU & Comb & Inform & Success & BLEU & Comb \\
        \midrule
        SimpleTOD & 84.40 & 70.10 & 15.01 & 92.26 & 85.00 & 70.50 & 15.23 & 92.98 \\
        $\text{UBAR}^\dag$ & 85.10 & 71.02 & 16.21 & 94.27 & 86.20 & 70.32 & 16.48 & 94.74 \\
        SOLOIST & 85.50 & 72.90 & 16.54 & 95.74 & - & - & - & - \\
        $\text{MinTL}^\dag$ & 84.88 & 74.91 & 17.89 & 97.78 & - & - & - & - \\
        TOP+NOD & 86.90 & 76.20 & 20.58 & 102.13 & - & - & - & - \\
        $\text{PPTOD}^\star$ & 89.20 & 79.40 & 18.62 & 102.92 & 87.09 & 79.08 & 19.17 & 102.26 \\
        MTTOD & 90.99 & 82.58 & 20.25 & 107.04 & 90.99 & 82.08 & 19.68 & 106.22 \\
        $\text{SPACE}^\star$ & 91.50 & 84.70 & 19.30 & 107.40 & \underline{93.00} & 84.10 & 19.91 & 108.46 \\
        \midrule
        {\begin{tabular}[l]{@{}c@{}}\ours{}\\[-1ex]{\tiny (\llamaSmall{})}\end{tabular}} & \textbf{94.30} & \textbf{87.10} & \textbf{21.48} & \textbf{112.18} & \textbf{94.40} & \textbf{87.50} & \textbf{21.41} & \textbf{112.38} \\
        \llamaSmall{} (FT) & \underline{92.80} & \underline{84.90} & \underline{20.95} & \underline{109.80} & 92.30 & \underline{84.60} & \underline{21.33} & \underline{109.78} \\
        \hdashline \\ [-2ex]
        {\begin{tabular}[l]{@{}c@{}}\ours{} (GT)\\[-1ex]{\tiny (\llamaSmall{})}\end{tabular}} & 88.20 & 81.60 & 21.87 & 106.77 & 87.60 & 81.10 & 21.92 & 106.27 \\
        \bottomrule
    \end{tabular*}
% \footnotetext{Best results are in bold, and the second best results are underlined.}
\footnotetext[\dag]{~During inference at current turn, corresponding methods require the orcale dialogue states of previous turns.}
\footnotetext[\star]{~Corresponding PCMs have been further fine-tuned on each dataset individually.}
\footnotetext{\hlyellow{\textsc{Llama 2 7B} (FT) signifies that \textsc{Llama 2 7B} model was fine-tuned exclusively on the training set of each dataset, without cross-dataset training.}}
\footnotetext{\hlyellow{\textsc{ESAinsTOD} (GT) indicates that results are based on ground-truth system actions and responses from previous turns.}}

% \footnotetext{When calculating evaluation metrics, the generated dialogue states are used to query the databases for DB results, which are then compared with the ground-truth DB results to obtain inform rate.}
\end{minipage}
\end{center}
\end{table}

\bmhead{End-to-end dialog modeling}
Tables~\ref{table:e2e_ck_full} and \ref{table:e2e_mwoz_full} present the end-to-end dialog modeling results on CamRest676, In-Car, and MultiWOZ datasets.
Our instruction-tuned model, \ours{}, outperforms previous state-of-the-art methods across all four datasets.
From Table~\ref{table:e2e_ck_full}, we can see that our model reaches the highest scores on all metrics for the CamRest676 and In-Car datasets.
For both MultiWOZ 2.0 and 2.1 datasets, \ours{} achieves absolute improvements of 4.78\% and 3.92\% in combined scores over the previous strongest baseline, SPACE, as shown in Table~\ref{table:e2e_mwoz_full}.
\ours{} also attains the highest SuccF1 or Success scores on In-Car and two MultiWOZ benchmarks, outperforming SPACE by 4.93, 2.40, and 3.40 points.
In particular, the BLEU score of \ours{} surpasses those of all baselines across all four benchmarks.
These advantages indicate that our dialog system can more accurately fulfill user requests and deliver smoother responses.
Notably, all compared baselines have been specifically fine-tuned on their respective datasets, with PPTOD and SPACE undergoing dialogue pre-training before fine-tuning.
In contrast, \ours{} is able to follow task instructions and schema information to adapt to various datasets that involve different domains and dialog tasks, making it more flexible and generalizable.

\begin{table}[t]
\begin{center}
\begin{minipage}{0.75\textwidth}
    \caption{DST evaluation: Joint Goal Accuracy on MultiWOZ}\label{table:dst_full}
    \begin{tabular*}{\linewidth}{@{\extracolsep{\fill}}lcc@{\extracolsep{\fill}}}
        \toprule
        \textbf{Model} & \textbf{MultiWOZ 2.0} & \textbf{MultiWOZ 2.1} \\
        \midrule \\ [-4ex]
        \multicolumn{3}{c}{\textit{Classification-based Approaches}} \\ [-1ex]
        \midrule
            TOD-BERT & - & 48.00 \\
            $\text{DST-Picklist}^\dag$ & \underline{54.39} & 53.30 \\
            $\text{SST}^\dag$ & 51.17 & 55.23 \\
            TripPy & - & 55.29 \\
            $\text{CHAN}^\dag$ & 52.68 & \underline{58.55} \\
            $\text{FPDSC-turn}^\dag$ & \textbf{55.03} & 57.88 \\
            $\text{FPDSC-dual}^\dag$ & 53.17 & \textbf{59.07} \\
        \midrule \\ [-4ex]
        \multicolumn{3}{c}{\textit{Hybrid Approaches}} \\ [-1ex]
        \midrule
            $\text{DSTQA}^\dag$ & 51.36 & 49.67 \\
            DS-DST & \underline{52.24} & \underline{51.21} \\
            $\text{DSS-DST}^\dag$ & \textbf{56.93} & \textbf{60.73} \\
        \midrule \\ [-4ex]
        \multicolumn{3}{c}{\textit{Generation-based Approaches}} \\ [-1ex]
        \midrule
            SOM-DST & 51.38 & 52.57 \\
            MinTL & 52.10 & 53.62 \\
            SimpleTOD & - & 55.76 \\
            Seq2seq-DU & - & 56.10 \\
            UBAR (DST) & 52.59 & 56.20 \\
            SOLOIST (DST) & 53.20 & 56.85 \\
            $\text{PPTOD}^\star$ (DST) & 53.89 & \underline{57.45} \\
            \midrule
            {\begin{tabular}[l]{@{}c@{}}\ours{}\\[-1ex]{\tiny (\llamaSmall{})}\end{tabular}} & \textbf{55.90} & \textbf{58.68} \\
            \llamaSmall{} (FT) & \underline{54.92} & 56.74 \\
            \hdashline \\ [-2ex]
            {\begin{tabular}[l]{@{}c@{}}\ours{} (GT)\\[-1ex]{\tiny (\llamaSmall{})}\end{tabular}} & 57.23 & 60.76 \\
        \bottomrule
    \end{tabular*}
% \footnotetext{Best results are in bold, and the second best results are underlined.}
\footnotetext{The end-to-end methods tagged with (DST) include versions specifically trained for dialogue state tracking (DST) task.}
\footnotetext[\dag]{~Corresponding methods necessitate a full ontology in which all possible values are predefined for each slot.}
\footnotetext[\star]{~Corresponding PCM has been further fine-tuned on each downstream dataset individually.}
\end{minipage}
\end{center}
\end{table}

\bmhead{Dialog state tracking}
The accuracy of tracking dialog states is a key determinant of end-to-end dialog modeling performance.
Therefore, we evaluate various methods for DST task, including both task-specific and end-to-end approaches, as detailed in Table~\ref{table:dst_full}.
SOM-DST, Seq2seq-DU, and all classification-based and hybrid approaches are specifically designed for DST, whereas the remaining methods are end-to-end.
Without relying on a complete ontology or any specialized post-processing, \ours{} obtains the highest joint goal accuracy (JGA) among all generation-based baselines.
Unlike other end-to-end methods that require specialized training to improve dialog state generation, our model consistently achieves superior results in both DST and response generation tasks within end-to-end dialog modeling.

In the MultiWOZ benchmarks, we adopt a more realistic setting where a TOD system can only access the dialog context containing its own generated content from previous turns, rather than relying on the ground-truth dialog states, system actions, and responses.
For a fair comparison, we also report DST results based on both generated dialogue states and ground-truth system actions and responses from previous turns, referred to as ``\ours{} (GT)'' in Table~\ref{table:dst_full}.
When given actual dialog history, our model performs comparably to, or even outperforms, state-of-the-art specialized methods such as FPDSC and DSS-DST in the DST task, highlighting its substantial potential for dialog understanding.
Similarly to the findings reported in UBAR, we observe that incorporating ground-truth responses into the dialog history leads to a significant decline in both the inform rate and success rate, as detailed in Table~\ref{table:e2e_mwoz_full}.
This decline occurs because \ours{} relies on its historical actions to make context-aware decisions for the current turn, and introducing ground-truth actions may disrupt the coherence of its dialog policy.

\bmhead{Language understanding}
We also report results on the test sets of the language understanding tasks included in our instruction-tuning corpus.
According to Tables~\ref{table:intent_full} and \ref{table:slu_full}, \ours{} demonstrates competitive performance compared to existing methods in intent detection and slot filling benchmarks, with the exception of SPACE.
It is worth mentioning that these intent detection datasets do not provide specific descriptions or explanations for abbreviated intent labels, which may make classification-based discriminative methods more appropriate.
Consequently, we suggest that such datasets may not be conducive to end-to-end modeling tasks and could negatively affect performance on other intent detection datasets.
Furthermore, we individually fine-tune \llamaSmall{} on each benchmark and present the evaluation results in the ``\llamaSmall{} (FT)'' row of Table~\ref{table:intent_full}.
Compared to another generation-based method, PPTOD, \llamaSmall{} (FT) achieves only a marginal improvement, indicating that merely increasing the model size does not yield significant benefits for intent detection tasks.
We attribute this phenomenon to the absence of natural language descriptions for intent labels, making it difficult for LLMs to align these labels with the user's natural language inputs.
This issue will be further explored in future work.

\begin{table}[t]
\begin{center}
\begin{minipage}{0.8\textwidth}
    \caption{Comparison results on intent detection tasks}
    \label{table:intent_full}
    \begin{tabular*}{\linewidth}{@{\extracolsep{\fill}}lccc@{\extracolsep{\fill}}}
        \toprule
        \textbf{Model} & \textbf{BANKING77} & \textbf{CLINIC150} & \textbf{HWU64} \\
        \midrule \\ [-4ex]
        \multicolumn{4}{c}{\textit{Classification-based Approaches}} \\ [-1ex]
        \midrule
            ConvBERT & 92.95 & 97.07 & 90.43 \\
            USE+ConvRT & 93.36 & 97.16 & 92.62 \\
            Example+Observer & 93.83 & 97.31 & \underline{93.03} \\
            ConvFiT & \underline{94.16} & \underline{97.34} & 92.42 \\
            $\text{SPACE}^\star$ & \textbf{94.94} & \textbf{97.89} & \textbf{94.14} \\
        \midrule \\ [-4ex]
        \multicolumn{4}{c}{\textit{Generation-based Approaches}} \\ [-1ex]
        \midrule
            $\text{PPTOD}^\star$ & \textbf{93.86} & \underline{97.13} & \textbf{92.84} \\
            \midrule
            {\begin{tabular}[l]{@{}c@{}}\ours{}\\[-1ex]{\tiny (\llamaSmall{})}\end{tabular}} & \underline{92.89} & \textbf{97.31} & \underline{92.75} \\
            \llamaSmall{} (FT) & 94.35 & 97.29 & 93.01 \\
        \bottomrule
    \end{tabular*}
\footnotetext{Best results are in bold, and the second best results are underlined.}
\footnotetext[\star]{~Corresponding PCMs have been further fine-tuned on each downstream dataset individually.}
\end{minipage}
\end{center}
\end{table}

\begin{table}[t]
\begin{center}
\begin{minipage}{0.9\textwidth}
    \caption{Comparison results on SNIPS dataset}
    \label{table:slu_full}
    \begin{tabular*}{\linewidth}{@{\extracolsep{\fill}}lccc@{\extracolsep{\fill}}}
        \toprule
        \textbf{Model} & Intent Acc & Slot F1 & Overall Acc \\
        \midrule
            BERT-Joint & \underline{99.00} & 96.20 & 91.60 \\
            \text{Stack-Propgation+BERT} & \underline{99.00} & \underline{97.00} & \underline{92.90} \\
            \text{Co-Interactive transformer+BERT} & 98.80 & \textbf{97.10} & \textbf{93.10} \\
            \midrule
            {\begin{tabular}[l]{@{}c@{}}\ours{}\\[-1ex]{\tiny (\llamaSmall{})}\end{tabular}} & \textbf{99.43} & 96.76 & 92.14 \\
        \bottomrule
    \end{tabular*}
\footnotetext{Best results are in bold, and the second best results are underlined.}
\end{minipage}
\end{center}
\end{table}

\begin{sidewaystable}
\sidewaystablefn
\begin{center}
\begin{minipage}{\textheight}
    \setlength{\tabcolsep}{1.6mm}
    \caption{Low-resource end-to-end evaluation on MultiWOZ 2.0}\label{table:e2e_few}
    \begin{tabular*}{\textheight}{@{\extracolsep{\fill}}lcccccccccccc@{\extracolsep{\fill}}}
        \toprule
        \multirow{2}{*}{\textbf{Model}} & \multicolumn{4}{@{}c@{}}{5\% of training data} & \multicolumn{4}{@{}c@{}}{10\% of training data} & \multicolumn{4}{@{}c@{}}{20\% of training data} \\ \cmidrule(lr){2-5} \cmidrule(lr){6-9} \cmidrule(lr){10-13}
        & Inform & Succ. & BLEU & Comb & Inform & Succ. & BLEU & Comb & Inform & Succ. & BLEU & Comb \\
        \midrule
        MD-Sequicity & 49.40 & 19.70 & 10.30 & 44.85 & 58.10 & 34.70 & 11.40 & 57.80 & 64.40 & 42.10 & 13.00 & 66.25 \\
        DAMD & 52.50 & 31.80 & 11.60 & 53.75 & 55.30 & 30.30 & 13.00 & 55.80 & 62.60 & 44.10 & 14.90 & 68.25 \\
        SOLOIST & 69.30 & 52.30 & 11.80 & 72.60 & 69.90 & 51.90 & 14.60 & 75.50 & 74.00 & 60.10 & 15.25 & 82.29 \\
        MinTL & 75.48 & 60.96 & 13.98 & 82.20 & 78.08 & 66.87 & 15.46 & 87.94 & 82.48 & 68.57 & 13.00 & 88.53 \\
        PPTOD & 79.86 & 63.48 & \underline{14.89} & 86.55 & 84.42 & 68.36 & \underline{15.57} & 91.96 & 84.94 & 71.70 & 17.01 & 95.32 \\
        \midrule
        {\begin{tabular}[l]{@{}c@{}}\ours{}\\[-1ex]{\tiny (\llamaSmall{})}\end{tabular}} & \textbf{88.70} & \textbf{79.10} & \textbf{16.14} & \textbf{100.04} & \textbf{90.30} & \textbf{82.20} & \textbf{16.37} & \textbf{102.62} & \textbf{91.80} & \textbf{84.20} & \textbf{20.69} & \textbf{108.69} \\
        \llamaSmall{} (FT) & \underline{86.00} & \underline{76.20} & 14.28 & \underline{95.38} & \underline{89.90} & \underline{80.70} & 15.15 & \underline{100.45} & \underline{89.20} & \underline{82.40} & \underline{18.21} & \underline{104.01} \\
        \bottomrule
    \end{tabular*}
\footnotetext{Best results are in bold, and the second best results are underlined.}
\footnotetext{Instruction-tuning data comprises the complete training sets of five end-to-end (E2E) dialog datasets, excluding CamRest676, along with corresponding proportions of the MultiWOZ 2.0 dataset.}
\footnotetext{Results of all baselines are taken from~\citet{DBLP:conf/acl/SuSMG0LZ22}. ``Succ.'' denotes the success rate.}
\end{minipage}
\end{center}
\end{sidewaystable}

\begin{table}[t]
\begin{center}
\begin{minipage}{0.65\textwidth}
    \setlength{\tabcolsep}{1.6mm}
    \caption{Low-resource DST evaluation on MultiWOZ 2.0}\label{table:dst_few}
    \begin{tabular*}{\linewidth}{@{\extracolsep{\fill}}lccccccc@{\extracolsep{\fill}}}
        \toprule
        \multirow{2}{*}{\textbf{Model}} & \multicolumn{3}{@{}c@{}}{\textbf{Training Percentage (\%)}} \\ \cmidrule{2-4}
            & 5 & 10 & 20 \\
        \midrule
        SimpleTOD & 16.14 & 22.37 & 31.22 \\
        MinTL & 21.28 & 30.32 & 35.96 \\
        SOLOIST & 26.53 & 32.42 & 38.68 \\
        $\text{PPTOD}_\text{base}$ & 40.20 & 43.45 & 46.96 \\
        $\text{PPTOD}_\text{large}$ & 43.61 & 45.96 & 48.95 \\
        \midrule
        {\begin{tabular}[l]{@{}c@{}}\ours{}\\[-1ex]{\tiny (\llamaSmall{})}\end{tabular}} & \textbf{50.41} & \textbf{51.48} & \textbf{52.89} \\
        \llamaSmall{} (FT) & \underline{47.69} & \underline{48.94} & \underline{51.26} \\ [+1ex]
        \hdashline \\ [-2ex]
        {\begin{tabular}[l]{@{}c@{}}\ours{} (GT)\\[-1ex]{\tiny (\llamaSmall{})}\end{tabular}} & 50.16 & \textbf{52.60} & \textbf{54.76} \\
        \llamaSmall{} (FT, GT) & 48.77 & 50.81 & 52.14 \\
        \bottomrule
    \end{tabular*}
% \footnotetext{Best results are in bold, and the second best results are underlined.}
\footnotetext{Instruction-tuning data comprises the complete training sets of five end-to-end (E2E) dialog datasets, excluding CamRest676, along with corresponding proportions of the MultiWOZ 2.0 dataset.}
\end{minipage}
\end{center}
\end{table}

% \subsection{Ablation study}

\subsection{Further analysis}
\label{sec:further_analysis}
To further explore the advantages of our instruction-tuning framework, we conduct a comprehensive study focusing on the following key questions:
\begin{itemize}
    % Large language models 是否适合直接应用于任务型对话建模？
    \item \textbf{\textit{Question 1}}: Are large language models (LLMs) suitable for direct application in task-oriented dialog modeling? (Refer to Section~\ref{sec:llm_strength})
    % \ours{}在端到端对话建模上的泛化能力如何？(low-resource and zero-shot)
    \item \textbf{\textit{Question 2}}: What is the generalization capability of \ours{} in end-to-end dialog modeling? (Refer to Section~\ref{sec:low_resource})
    % schema information 是否可以提升模型的data-efficiency？
    \item \textbf{\textit{Question 3}}: Can schema information improve the data efficiency of the model? (Refer to Section~\ref{sec:data_efficiency})
    % dialog context 对 session-level dialog modeling 有何影响？
    \item \textbf{\textit{Question 4}}: What impact does dialog context have on session-level dialog modeling? (Refer to Section~\ref{sec:dialog_context})
    % schema information 是否可以降低end-to-end对话中的级联错误？
    \item \textbf{\textit{Question 5}}: Can schema information reduce cascading errors in end-to-end dialog modeling? (Refer to Section~\ref{sec:dialog_context})
\end{itemize}

\subsubsection{Explore the strengths of large language models}\label{sec:llm_strength}
% \ours{} leverages large language models (LLMs) that have demonstrated capabilities matching or even surpassing human-level performance across numerous NLP tasks, showcasing exceptional instruction adherence and conversational proficiency.
% To assess the applicability of LLMs in TOD modeling, \llamaSmall{} is directly fine-tuned on the MultiWOZ dataset under both full-resource and low-resource settings.
\hlyellow{To explore \textit{``Question 1: Are large language models (LLMs) suitable for direct application in task-oriented dialog modeling?''}, we first assess the base applicability of LLMs.}
\ours{} leverages LLMs that have demonstrated exceptional instruction adherence and conversational proficiency.
To determine their suitability for TOD modeling, we directly fine-tune \llamaSmall{} on the MultiWOZ dataset under both full-resource and low-resource settings.

The resulting model, referred to as ``\llamaSmall{} (FT)'', consistently outperforms all baselines, including the pre-trained conversational models PPTOD and SPACE, as detailed in Tables~\ref{table:e2e_mwoz_full} and \ref{table:dst_full}.
In low-resource setting, we used only 5\%, 10\%, and 20\% of the training data for fine-tuning. 
The results, as listed in the ``\llamaSmall{} (FT)'' row of Tables~\ref{table:e2e_few} and \ref{table:dst_few}, show that \llamaSmall{} achieves substantial advantages in terms of generalization over PPTOD across different data proportions.
To be specific, \llamaSmall{}, utilizing just 5\% of the training data, attains a combined score of 95.38, which is equivalent to PPTOD's 95.32 achieved with 20\% of the training data. Moreover, \llamaSmall{} with 20\% of the training data achieves a score of 104.01, surpassing PPTOD trained on the entire dataset, which scored 102.92.
These results reveal the potential of LLMs in transfer learning for task-oriented dialog modeling and imply that continuous dialogue pre-training may not be necessary for such models.

\subsubsection{Analysis of generalization capability in end-to-end (E2E) dialog modeling}\label{sec:low_resource}
\hlyellow{This subsection addresses \textit{``Question 2: What is the generalization capability of \textsc{ESAinsTOD} in end-to-end dialog modeling?''}.}
As discussed in Section~\ref{sec:instruction_schema_alignment}, our instruction-tuning method is designed to adapt flexibly to various dialog tasks and domains by leveraging task instructions and schema information.
To verify the effectiveness and generalization power of our approach, we conduct both low-resource and zero-shot end-to-end evaluations on the MultiWOZ datasets.
\bmhead{Low-resource evaluation}
The model is fine-tuned on the training sets of five other task-oriented dialog datasets, supplemented with varying proportions (i.e., 5\%, 10\%, 20\%) of the MultiWOZ 2.0 dataset, and then evaluated on the MultiWOZ 2.0 test set.
We compare \ours{} with several state-of-the-art methods, including MD-Sequicity~\citep{DBLP:conf/aaai/ZhangOY20}, DAMD~\citep{DBLP:conf/aaai/ZhangOY20}, SOLOIST, MinTL, and PPTOD.
The experimental results, presented in Tables~\ref{table:e2e_few} and \ref{table:dst_few}, indicate that \ours{} consistently outperforms all baselines across different amounts of MultiWOZ 2.0 training data.
Notably, \ours{} surpasses PPTOD by significant margins in terms of Inform and Success rates, particularly when only 5\% of the training data is available (i.e., 88.70\% vs. 79.86\% in Inform rate, 79.10\% vs. 63.48\% in Success rate).
This performance underscores \ours{}'s capability to maintain robust dialogue modeling and understanding even in low-resource settings.
Additionally, \ours{} slightly outperforms \llamaSmall{} (FT), meaning that our instruction-tuning framework can efficiently integrate diverse dialog datasets with varied annotation schemas, domains, and tasks.
This integration facilitates the transfer of dialog-related knowledge, such as general dialog policies, resulting in further performance enhancements in both full-resource and low-resource settings.
When trained on 20\% of the MultiWOZ 2.0 data, our \ours{} is even comparable to the previous state-of-the-art baseline SPACE with full data, as detailed in Table~\ref{table:e2e_mwoz_full}.

\bmhead{Zero-shot evaluation}

\begin{table}[t]
\begin{center}
\begin{minipage}{0.9\textwidth}
    \setlength{\tabcolsep}{1.6mm}
    \caption{Zero-shot evaluation results under different ablation settings}\label{table:zero_shot}
    \begin{tabular*}{\linewidth}{@{\extracolsep{\fill}}lccccc@{\extracolsep{\fill}}}
        \toprule
        \textbf{Model} & JGA & Inform & Success & BLEU & Combined \\
        % \midrule \\ [-4ex]
        % \multicolumn{6}{c}{\textit{\textbf{MultiWOZ 2.0}}} \\ [-1ex]
        % \midrule
        % {\begin{tabular}[l]{@{}c@{}}\ours{}\\[-1ex]{\tiny (\llamaSmall{})}\end{tabular}} & \textbf{50.14} & \textbf{88.20} & \textbf{47.10} & 6.25 & \textbf{73.90} \\
        % \hdashline \\ [-2ex]
        % \qquad\textit{w/o sa} & 44.15 & 60.20 & 31.40 & \textbf{6.70} & 52.50 \\
        % \qquad\textit{w/o ia, sa} & 41.17 & 60.70 & 28.20 & 5.57 & 50.02 \\
        % \midrule \\ [-4ex]
        % \multicolumn{6}{c}{\textbf{\textit{MultiWOZ 2.1}}} \\ [-1ex]
        % \midrule
        % {\begin{tabular}[l]{@{}c@{}}\ours{}\\[-1ex]{\tiny (\llamaSmall{})}\end{tabular}} & \textbf{50.73} & \textbf{87.60} & \textbf{47.20} & 6.22 & \textbf{73.62} \\
        % \hdashline \\ [-2ex]
        % \qquad\textit{w/o sa} & 42.86 & 60.70 & 32.70 & \textbf{6.74} & 53.44 \\
        % \qquad\textit{w/o ia, sa} & 40.67 & 62.00 & 29.00 & 5.57 & 51.07 \\

        \midrule \\ [-4ex]
        \multicolumn{6}{c}{\textit{Turn-level evaluation on MultiWOZ 2.1 test set}} \\ [-1ex]
        \midrule
        {\begin{tabular}[l]{@{}c@{}}\ours{}\\[-1ex]{\tiny (\llamaSmall{})}\end{tabular}} & \textbf{63.00} & \textbf{91.60} & \textbf{73.50} & 6.60 & \textbf{89.15} \\
        \hdashline \\ [-2ex]
        \qquad\textit{w/o ia} & 56.17 & 87.80 & 67.00 & 6.61 & 84.01 \\
        \qquad\textit{w/o sa} & 48.67 & 69.20 & 39.30 & \textbf{6.88} & 61.13 \\
        \qquad\textit{w/o ia, sa} & 43.38 & 67.30 & 31.70 & 6.65 & 56.15 \\
        \midrule \\ [-4ex]
        \multicolumn{6}{c}{\textit{Session-level evaluation on MultiWOZ 2.1 test set}} \\ [-1ex]
        \midrule
        {\begin{tabular}[l]{@{}c@{}}\ours{}\\[-1ex]{\tiny (\llamaSmall{})}\end{tabular}} & \textbf{26.90} & \textbf{86.00} & \textbf{63.30} & \textbf{6.67} & \textbf{81.32} \\
        \hdashline \\ [-2ex]
        \qquad\textit{w/o ia} & 14.84 & 62.30 & 40.20 & 6.49 & 57.74 \\
        \qquad\textit{w/o sa} & 7.14 & 28.70 & 13.20 & 6.63  & 27.58 \\
        \qquad\textit{w/o ia, sa} & 3.45 & 23.30 & 9.50 & 5.94 & 22.34 \\
        \bottomrule
    \end{tabular*}
\footnotetext{For turn-level evaluation, where oracle historical dialog states are included in the dialog context, the Joint Goal Accuracy (JGA) primarily measures the ability of the dialogue system to recognize the user's needs in the current turn.}
\end{minipage}
\end{center}
\end{table}

To assess the adaptability of our proposed instruction-tuning framework in unseen dialog scenarios, we fine-tune \llamaSmall{} on five TOD datasets before applying it directly to the MultiWOZ test set. 
In particular, in addition to the conventional session-level evaluation, we adopt a turn-level evaluation setting in which intermediate task results from historical turns in the dialog context are considered as ground truth.
This evaluation mode concentrates on assessing the system's comprehension of the dialogue context as well as its ability to complete tasks at each turn.
Table~\ref{table:zero_shot} presents the zero-shot evaluation results under different fine-tuning settings.
% (91.60-69.20 + 86.00-28.70) / 2 = 39.85% 和 (73.50-39.30 + 63.30-13.20) / 2 = 42.15%
When the schema-aware mechanism is removed from the instruction-tuning process, the new model (denoted as \textit{w/o sa}) experiences a substantial reduction in performance, with an average decrease of 39.85\% in the Inform rate and 42.15\% in the Success rate across both evaluation settings.
This emphasizes the crucial role of aligning dialogue schemas with task outputs in enhancing the model's generalization capabilities.
Furthermore, removing task instructions from the fine-tuning data (denoted as \textit{w/o ia}) leads to significant declines in four out of five metrics, implying the importance of task instructions in distinguishing diverse task flows within different dialog scenarios.
Compared to models fine-tuned with only one alignment mechanism (either \textit{w/o ia} or \textit{w/o sa}), the model without any alignment mechanism (denoted as \textit{w/o ia, sa}) suffers more severe performance degradation.
From the opposite perspective, employing both schema-aware and instruction-aware mechanisms simultaneously enhances the generalization capabilities of the dialog system, ensuring optimal zero-shot performance on new dialogue schemas and task flows.

Both low-resource and zero-shot evaluation results affirm the effectiveness of our proposed framework in low-resource task-oriented dialog modeling and emphasize the critical role of these two alignment mechanisms in improving the performance of dialog systems in unseen dialog scenarios.

\subsubsection{Data-efficiency analysis}\label{sec:data_efficiency}
\begin{figure*}[t]
    \centering
    \subfloat[Inform Rate on MultiWOZ 2.1]{\includegraphics[width=0.48\textwidth]{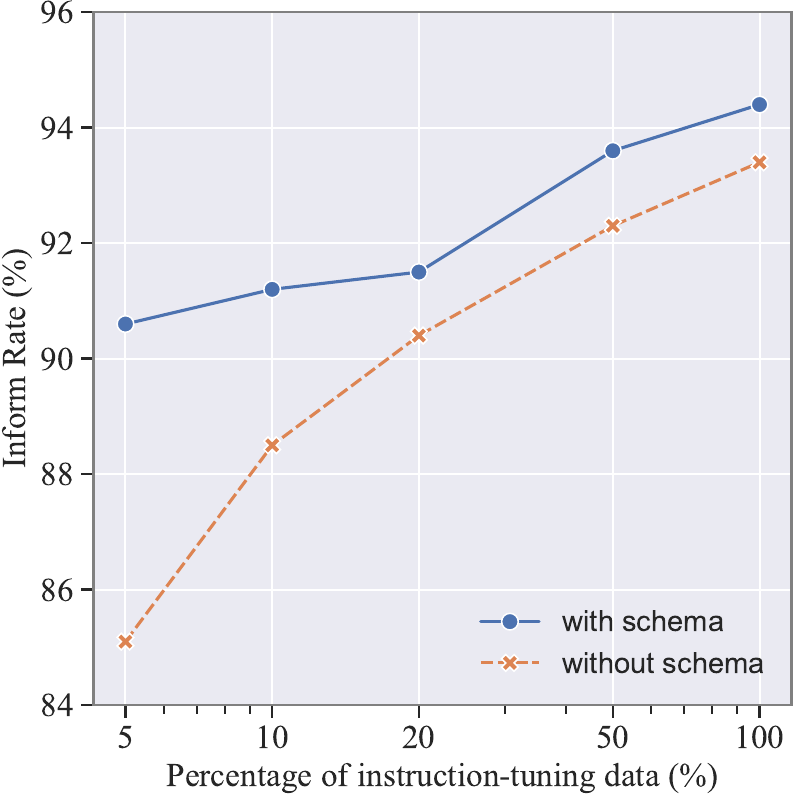}\label{fig:inform_rate_on_varying_percentages}}\hspace{0.02\textwidth}
    \subfloat[Success Rate on MultiWOZ 2.1]{\includegraphics[width=0.48\textwidth]{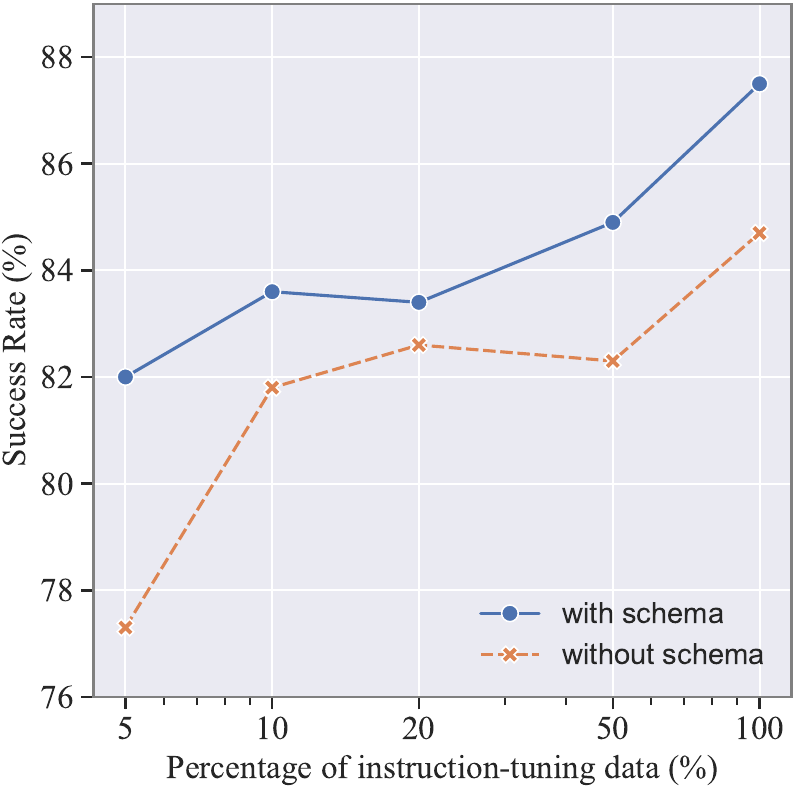}\label{fig:success_rate_on_varying_percentages}}
    \caption{Evaluation results of \ours{} trained on varying percentages of our instruction-tuning corpus with or without schema information}
    \label{fig:metrics_on_varying_percentages}
\end{figure*}
% experiments on data scales for proving the data-efficiency of the model
% zero-shot evaluation for proving the transferability of the model

% To investigate the impact of schema information on the data efficiency of ESAinsTOD in end-to-end dialog modeling, we conduct a data-scaling experiment on the MultiWOZ 2.1 dataset.
\hlyellow{To answer \textit{``Question 3: Can schema information improve the data efficiency of the model in end-to-end dialog modeling?''}, we conduct a data-scaling experiment.}
Concretely, we first randomly sample certain percentages of training data (i.e., 5\%, 10\%, 20\%, and 50\%) from seven E2E dialog datasets and then mix them together to construct the instruction-tuning data for fine-tuning \llamaSmall{}.
The ablation study in Fig.~\ref{fig:metrics_on_varying_percentages} illustrates how dataset-specific schema information influences task completion of our task-oriented dialog system.

With limited training data, particularly at 5\% and 10\%, \ours{} demonstrates improved performance by delivering more relevant entities and responding more accurately to user requests, compared to our implemented baseline that lacks schema information.
This enhancement is quantitatively reflected in higher Inform and Success rates, underscoring the critical role of schema information in boosting the model's generalization capabilities in extremely low-resource scenarios.
By incorporating schema information, the model can effectively overlook the discrepancies between different datasets and learn more generalized dialogue policies.
As the scale of training data increases (i.e., 20\% and 50\%), the performance gains from schema information gradually diminish. 
We speculate that with more data, the model effectively discerns the unique characteristics of each dataset, including the language styles of user and system utterances (e.g., length, case), as well as dataset-specific features (e.g., domain names, task names).
Consequently, this comprehensive understanding reduces the relative importance of schema information.
Nonetheless, our \ours{} still achieves a higher completion rate even with 100\% of the training data, indicating that schema alignment can further elevate the performance ceiling of end-to-end task-oriented dialog modeling.

To conclude, the integration of schemas enables the model to achieve comparable performance with less training data, thereby enhancing the data efficiency of the model.

\subsubsection{Influence of dialog context}\label{sec:dialog_context}

\begin{figure*}[t]
    \centering
    \includegraphics[width=1.0\textwidth]{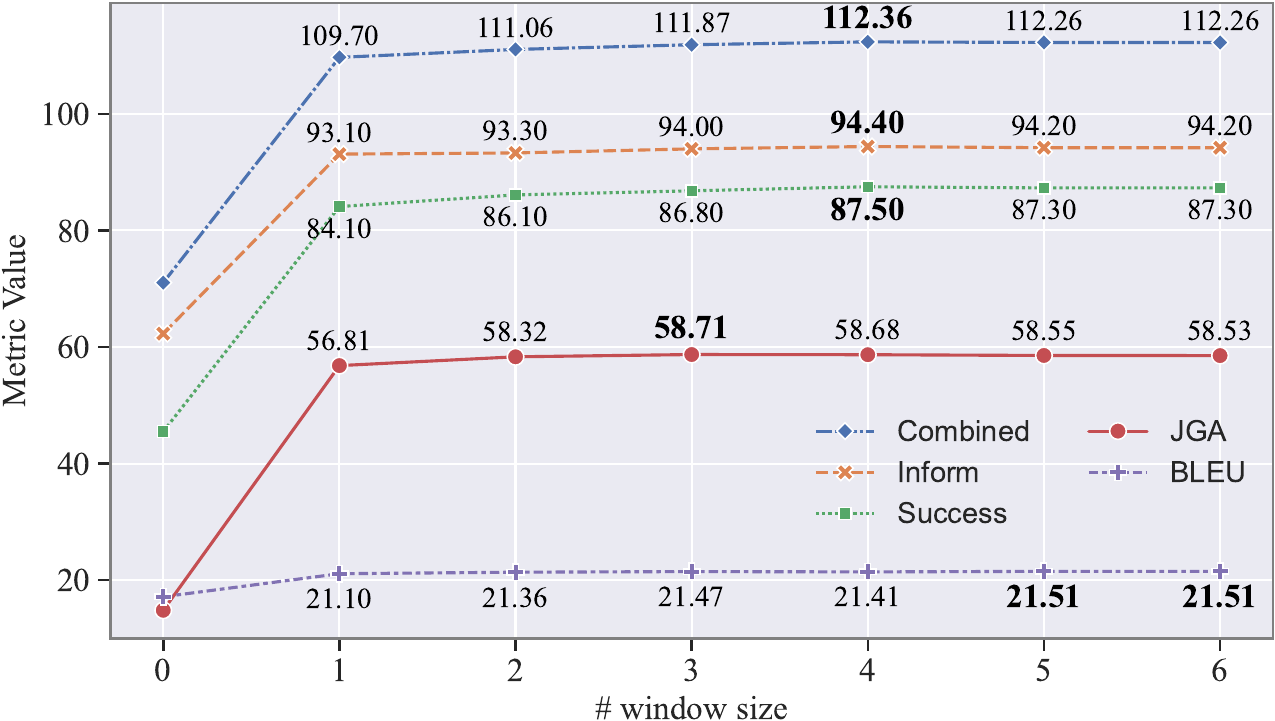}
    \caption{Results of \ours{} on MultiWOZ 2.1 with varying context window sizes}
    \label{fig:context_window}
\end{figure*}

Instead of the traditional turn-level approach, \ours{} is fine-tuned and evaluated in a session-level end-to-end manner.
Specifically, the outputs of various tasks in previous turns are retained in the dialog context and may influence task predictions in subsequent turns.
% Therefore, we conduct several experiments to investigate the impact of dialog context on session-level dialog modeling.
\hlyellow{Therefore, we conduct several experiments to investigate two related questions: \textit{``Question 4: What impact does dialog context have on session-level dialog modeling?''} and \textit{``Question 5: Can schema information reduce cascading errors in end-to-end dialog modeling?''}.}

\bmhead{Context window size}
We first restrict the number of dialog history turns that \ours{} can access when inferring each dialog turn to investigate the relationship between the context window size and inference results, as shown in Fig.~\ref{fig:context_window}.
As the context window increases, the results on MultiWOZ 2.1 show a trend of rapid growth followed by a slow decline, with the combined score peaking when the context window is 4.
It is evident that sufficient dialog history is necessary for the model to plan reasonable and coherent dialog behaviors.
However, the redundancy of excessive dialog context may slightly interfere with state tracking and decision-making, signifying opportunities for future research.

\begin{table}[t]
\begin{center}
\begin{minipage}{1.0\textwidth}
    \setlength{\tabcolsep}{1mm}
    \caption{Session-level cascading error analysis of \ours{} on MultiWOZ 2.1}\label{table:session_cascade}
    \begin{tabular*}{\linewidth}{@{\extracolsep{\fill}}cccllll@{\extracolsep{\fill}}}
        \toprule
        \multirow{2}{*}{\textbf{Schema}} & \multicolumn{1}{@{}c@{}}{\textbf{Context}} & \multicolumn{1}{@{}c@{}}{\textbf{Current}} & \multirow{2}{*}{Inform} & \multirow{2}{*}{Success} & \multirow{2}{*}{BLEU} & \multirow{2}{*}{Combined} \\ \cmidrule(lr){2-2} \cmidrule(lr){3-3}
        & Belief & Belief & & & & \\
        \midrule
        $\times$ & GT & Gen & 94.50 & 85.80 & 19.61 & 109.76 \\
        $\times$ & Gen & Gen & 93.40  ($-$1.1) & 84.70 ($-$1.1) & 19.54 ($-$0.07) & 108.59 ($-$1.17) \\
        \hdashline \\ [-2ex]
        \checkmark & GT & Gen & 94.50 & 87.80 & 21.33 & 112.48 \\
        \checkmark & Gen & Gen & 94.40 ($-$\textbf{0.1}) & 87.50 ($-$\textbf{0.3}) & 21.41 ($+$\textbf{0.08}) & 112.36 ($-$\textbf{0.12}) \\
        \bottomrule
    \end{tabular*}
\footnotetext{We report the inference results conditioned on dialogue history turns that contain either ground-truth or generated dialogue states. The numbers in parentheses denote the cascading impact of the model's predictions from previous turns on the overall performance of dialogue modeling.}
\end{minipage}
\end{center}
\end{table}

\bmhead{Cascade error analysis}
Cascading error propagation is prevalent among various components of TOD systems and serves as a major obstacle to the robustness of dialog systems.
Next, by controlling the source of dialog states, we analyze how our proposed instruction-tuning framework alleviates session-level and task-level cascading errors in end-to-end dialog modeling.
Table~\ref{table:session_cascade} shows that the inference results of \ours{} on MultiWOZ 2.1 would decrease when the dialog context includes generated dialogue states from previous turns, compared to using ground-truth dialog states.
However, the introduction of schema information significantly mitigates this performance degradation, with a notably smaller reduction in the combined score ($-$1.17 vs. $-$0.12), which underscores enhanced system stability.
We attribute this to the fact that aligning the dialog-specific schema with the task outputs helps avoid interference from other dialogues in similar domains but with different schemas during fine-tuning, thereby facilitating more fine-grained dialog modeling.
Additionally, a more interesting phenomenon emerges in the task-level cascading error analysis, where the database query, dialog policy planning, and response generation for each dialog turn are based on ground-truth dialog states of the current turn.
As illustrated in Table~\ref{table:task_cascade}, under both settings — whether or not ground-truth previous dialog states are included in the dialog context — we observe that the use of schema information allows the model to achieve higher Inform and Success rates with the generated dialog states of the current turn than using the ground-truth dialog states.
In contrast, models without schema information exhibits significant cascading error propagation.
This finding suggests that our instruction-tuning framework offers considerable robustness against annotation errors found in the MultiWOZ 2.1 dataset, highlighting a notable level of resistance to data noise.

\begin{table}[t]
\begin{center}
\begin{minipage}{1.0\textwidth}
    \setlength{\tabcolsep}{1mm}
    \caption{Task-level cascading error analysis of \ours{} on MultiWOZ 2.1}\label{table:task_cascade}
    \begin{tabular*}{\linewidth}{@{\extracolsep{\fill}}cccllll@{\extracolsep{\fill}}}
        \toprule
        \multirow{2}{*}{\textbf{Schema}} & \multicolumn{1}{@{}c@{}}{\textbf{Context}} & \multicolumn{1}{@{}c@{}}{\textbf{Current}} & \multirow{2}{*}{Inform} & \multirow{2}{*}{Success} & \multirow{2}{*}{BLEU} & \multirow{2}{*}{Combined} \\ \cmidrule(lr){2-2} \cmidrule(lr){3-3}
        & Belief & Belief & & & & \\
        \midrule
        $\times$ & Gen & GT & 94.80 & 86.10 & 19.70 & 110.15 \\
        $\times$ & Gen & Gen & 93.40 ($-$1.4) & 84.70 ($-$1.4) & 19.54 ($-$0.16) & 108.59 ($-$1.56) \\
        \hdashline \\ [-2ex]
        \checkmark & Gen & GT & 94.10 & 86.90 & 21.52 & 112.02 \\
        \checkmark & Gen & Gen & 94.40 ($+$\textbf{0.3}) & 87.50 ($+$\textbf{0.6}) & 21.41 ($-$\textbf{0.11}) & 112.36 ($+$\textbf{0.34}) \\
        \midrule
        $\times$ & GT & GT  & 94.70 & 86.40 & 19.76 & 110.31 \\
        $\times$ & GT & Gen & 94.50 ($-$0.2) & 85.80 ($-$0.6) & 19.61 ($-$0.15) & 109.76 ($-$0.55) \\
        \hdashline \\ [-2ex]
        \checkmark & GT & GT & 93.90 & 86.70 & 21.57 & 111.87 \\
        \checkmark & GT & Gen & 94.50 ($+$\textbf{0.6}) & 87.80 ($+$\textbf{1.1}) & 21.33 ($-$\textbf{0.24}) & 112.48 ($+$\textbf{0.61}) \\
        \bottomrule
    \end{tabular*}
\end{minipage}
\end{center}
\end{table}

\subsection{\sechl{Ablation study}}
\label{sec:ablation_study}
In this section, to rigorously evaluate the contributions of our proposed components and dissect their individual impacts, we conduct a series of ablation studies.
These experiments are designed to investigate: (1) the effectiveness of our schema and instruction alignment mechanisms in a fair and controlled setting; (2) the influence of backbone model scale on overall performance; and (3) the efficiency gains afforded by our schema management algorithm.

\subsubsection{Effectiveness of alignment mechanisms}\label{sec:ablation_mechanisms}

A primary concern is to isolate the benefits of our methodology from the influence of the backbone model.
To ensure a fair comparison, we conduct a direct head-to-head evaluation where two representative baselines are fine-tuned on the same Qwen2.5 backbone models and the same unified corpus as our proposed \ours{}.
Furthermore, to precisely quantify the importance of each alignment mechanism, we compare our full \ours{} model against two ablated variants.
All models were trained on the unified corpus and evaluated on the MultiWOZ 2.1 test set.
The selected baselines are as follows:
\begin{itemize}
\item \textbf{\ours{} \textit{w/o ia}} is trained without instruction alignment.
\item \textbf{\ours{} \textit{w/o sa}} is trained without schema alignment.
\item \textbf{UBAR}~\citep{DBLP:conf/aaai/YangLQ21}, which involves session-level dialogue modeling and is functionally equivalent to our framework but without the schema and instruction alignment mechanisms.
\item \textbf{PPTOD}~\citep{DBLP:conf/acl/SuSMG0LZ22}, which adopts a multi-task learning paradigm by decomposing dialogue data into (task prompt, dialog input, task output) triplets. This method does not explicitly model the relationships between different dialogue tasks within a turn.
\end{itemize}

The results, presented in Table~\ref{tab:ablation_mechanisms}, lead to several key insights.
First, \ours{} consistently outperforms all baselines and ablated variants across all model sizes, demonstrating the synergistic benefits of our complete framework.
The significant performance delta between \ours{} and UBAR directly validates the efficacy of our proposed alignment mechanisms, as this is the primary architectural difference between them.

Second, the results allow us to dissect the distinct contributions of each component.
Schema alignment generally has a more substantial impact on performance, particularly for dialogue state tracking.
For instance, its removal (\textbf{\textit{w/o sa}}) results in the poorest state tracking performance among the ablated variants.
Across most model sizes, schema alignment contributes more to the Inform and Success rates than instruction alignment.
These findings suggest that schema alignment provides the model with a generalizable ability to follow structured schema definitions, while instruction alignment offers the flexibility to adapt to diverse dialogue task flows.

Finally, \ours{}'s superiority over PPTOD highlights the advantage of our session-level modeling approach, which better captures the complex inter-task dependencies inherent in realistic, multi-domain conversations.

\begin{table}[t]
\begin{center}
\begin{minipage}{1.0\textwidth}
    \setlength{\tabcolsep}{1mm}
    \caption{Ablation study on MultiWOZ 2.1 using Qwen2.5 model family}\label{tab:ablation_mechanisms}
    \begin{tabular*}{\linewidth}{@{\extracolsep{\fill}}lcccccc@{\extracolsep{\fill}}}
        \toprule
        \textbf{Method} & \textbf{Session-Level} & JGA & Inform & Success & BLEU & Comb \\
        \midrule \\ [-4ex]
        \multicolumn{7}{@{}l@{}}{\textbf{\textit{Backbone model: Qwen2.5-0.5B-Instruct}}} \\
        UBAR & \checkmark & 55.17 & 86.80 & 78.40 & 18.37 & 100.97 \\
        PPTOD & $\times$ & 51.22 & 89.90 & 81.30 & 19.78 & 105.38 \\
        \cdashline{1-7}[1.5pt/3pt] \\ [-2ex]
        \ours{} & \checkmark & \textbf{56.43} & \textbf{93.40} & \textbf{85.30} & 19.56 & \textbf{108.91} \\
        \qquad\textit{w/o ia} & \checkmark & 53.68 & 88.20 & 77.50 & \textbf{20.07} & 102.92 \\
        \qquad\textit{w/o sa} & \checkmark & 52.82 & 88.70 & 80.90 & 19.41 & 104.21 \\
        \midrule \\ [-4ex]
        
        \multicolumn{7}{@{}l@{}}{\textbf{\textit{Backbone model: Qwen2.5-1.5B-Instruct}}} \\
        UBAR & \checkmark & 55.40 & 85.90 & 77.40 & 19.44 & 101.09 \\
        PPTOD & $\times$ & 53.73 & 91.10 & 82.30 & 19.88 & 106.58 \\
        \cdashline{1-7}[1.5pt/3pt] \\ [-2ex]
        \ours{} & \checkmark & \textbf{57.30} & \textbf{93.40} & \textbf{85.10} & \textbf{20.06} & \textbf{109.31} \\
        \qquad\textit{w/o ia} & \checkmark & 56.58 & 92.30 & 83.00 & 19.79 & 107.44 \\
        \qquad\textit{w/o sa} & \checkmark & 56.19 & 86.90 & 79.30 & 19.85 & 102.95 \\
        \midrule \\ [-4ex]

        \multicolumn{7}{@{}l@{}}{\textbf{\textit{Backbone model: Qwen2.5-3B-Instruct}}} \\
        UBAR & \checkmark & 56.08 & 86.20 & 78.40 & 18.95 & 101.25 \\
        PPTOD & $\times$ & 54.19 & 92.00 & 83.50 & 19.80 & 107.55 \\
        \cdashline{1-7}[1.5pt/3pt] \\ [-2ex]
        \ours{} & \checkmark & \textbf{57.58} & \textbf{94.70} & \textbf{86.70} & \textbf{19.90} & \textbf{110.60} \\
        \qquad\textit{w/o ia} & \checkmark & 56.84 & 93.30 & 81.90 & 19.39 & 106.99 \\
        \qquad\textit{w/o sa} & \checkmark & 53.53 & 87.20 & 80.00 & 19.45 & 103.05 \\
        \bottomrule
    \end{tabular*}
\footnotetext{\textbf{Session-Level} stands for session-level end-to-end task-oriented dialog modeling.}
\end{minipage}
\end{center}
\end{table}

\subsubsection{Effect of backbone model scale}\label{sec:ablation_model_scale}

An important question is how model scale influences the capacity of different TOD methodologies to perform complex dialogue tasks.
To investigate this, we evaluate UBAR, PPTOD, and \ours{} using various sizes of the Qwen2.5 series as the backbone.

The results are also summarized in Table~\ref{tab:ablation_mechanisms}.
The performance of the baseline UBAR shows almost no improvement as the model size increases, suggesting that simply scaling up the model without a robust methodological framework yields diminishing returns.
In contrast, both PPTOD and our \ours{} exhibit moderate and consistent performance gains with larger backbone models.

This finding leads to a crucial insight: \textbf{\textit{merely increasing the size and power of the backbone LLM is insufficient to guarantee improved task-oriented dialogue capabilities}}.
Instead, it is the methodological design—such as the alignment mechanisms in \ours{} and the multi-task formulation in PPTOD—that is key to effectively unlocking and leveraging the inherent potential of the underlying language models.
Our approach demonstrates superior scaling properties, indicating its effectiveness in harnessing the power of larger models.

\subsubsection{Impact of schema management}\label{sec:ablation_management}

Next, we present further empirical analyses on the schema management strategy introduced in Algorithm~\ref{alg:schema_management}.
Algorithm~\ref{alg:schema_management} aims to enhance both training and inference efficiency by managing the repetition of schemas.
We quantify its impact from two perspectives.

\begin{table}[t]
\begin{center}
\begin{minipage}{1.0\textwidth}
    \setlength{\tabcolsep}{1mm}
    \caption{Impact of schema management on the statistics of our unified instruction-tuning dataset}\label{tab:schema_management_stats}
    \begin{tabular*}{\linewidth}{@{\extracolsep{\fill}}ccccc@{\extracolsep{\fill}}}
        \toprule
        \makecell{\textbf{Schema}\\\textbf{Management}} & Samples $\downarrow$ & Turns per Sample $\uparrow$ & Tokens per Turn $\downarrow$ & \makecell{Total Tokens\\(Billion)} $\downarrow$ \\
        \midrule
        $\times$ & 63,008 & 7.53 & 401.7 & 0.184 \\
        \checkmark & \textbf{47,489} & \textbf{8.63} & \textbf{252.8} & \textbf{0.103} \\
        \bottomrule
    \end{tabular*}
\footnotetext{Tokens are counted by the Qwen2.5 tokenizer. $\uparrow$: higher is better and $\downarrow$: lower is better.}
\footnotetext{The maximum token length per sample is 4096; dialogues exceeding this limit are split into multiple samples (see Section~\ref{sec:e2e_ft_and_eval} for details).}
\end{minipage}
\end{center}
\end{table}

First, on training data volume, Algorithm~\ref{alg:schema_management} avoids redundant schema definitions within training samples, significantly reducing the total number of tokens required for training.
As detailed in Table~\ref{tab:schema_management_stats}, applying this algorithm reduces the total token count of our unified corpus by approximately 44\%, from 0.184 billion to 0.103 billion, thereby lowering computational costs.
This management strategy also enhances the scope and training efficiency of session-level dialogue modeling by increasing the average number of turns per sample from 7.53 to 8.63.
By reducing the average tokens per turn, more conversational context can be accommodated within a single training instance, broadening the model's effective receptive field.

\begin{table}[t]
\begin{center}
\begin{minipage}{1.0\textwidth}
    \setlength{\tabcolsep}{1mm}
    \caption{Runtime efficiency comparison on MultiWOZ 2.1 using Qwen2.5-0.5B-Instruct as the backbone}\label{tab:inference_efficiency}
    \begin{tabular*}{\linewidth}{@{\extracolsep{\fill}}lcccccccc@{\extracolsep{\fill}}}
        \toprule
        \multirow{2}{*}{\textbf{Method}} & \multirow{2}{*}{\textbf{SM}} & \multicolumn{5}{@{}c@{}}{\textbf{End-to-End Dialogue Modeling}} & \multicolumn{2}{@{}c@{}}{\textbf{Inference Measurement}} \\ \cmidrule(lr){3-7} \cmidrule(lr){8-9}
        && JGA & Inform & Success & BLEU & Comb & Latency (ms) $\downarrow$ & Speedup $\uparrow$ \\
        \midrule
        PPTOD & - & 51.22 & 89.90 & 81.30 & \textbf{19.78} & 105.38 & \textbf{20.94} & \textbf{1.00$\times$} \\
        UBAR & - & 55.17 & 86.80 & 78.40 & 18.37 & 100.97 & 25.96 & 0.81$\times$ \\
        \midrule
        \multirow{2}{*}{\ours{}} & $\times$ & \textbf{56.52} & \textbf{93.50} & 85.00 & 19.54 & 108.79 & 40.59 & 0.52$\times$ \\
        & \checkmark & 56.43 & 93.40 & \textbf{85.30} & 19.56 & \textbf{108.91} & 30.65 & 0.68$\times$ \\
        \bottomrule
    \end{tabular*}
\footnotetext{\textbf{SM} stands for Schema Management. $\uparrow$: higher is better and $\downarrow$: lower is better.}
\footnotetext{The latency of each method is measured on a single Nvidia A100 80GB GPU using offline inference with vLLM and is calculated by dividing the total inference time by the total number of dialogue turns.}
\end{minipage}
\end{center}
\end{table}

Second, on inference performance, we report the end-to-end dialogue modeling results and average inference latency on the MultiWOZ 2.1 test set.
As shown in Table~\ref{tab:inference_efficiency}, applying schema management provides \ours{} with a notable reduction in inference latency (from 40.59 ms to 30.65 ms), corresponding to a 1.32x speedup.
This is primarily because eliminating repeated schema definitions reduces the prefilling latency during inference.
Crucially, this efficiency gain is achieved with no degradation in task performance metrics, as evidenced by a marginal difference of 0.12 in combined scores between configurations with and without schema management.
This result further underscores the robustness of \ours{} in handling contextual redundancy.

Moreover, we compare our \ours{} with two strong baselines, PPTOD and UBAR.
Although our method incurs a slightly higher latency (30.65 ms) than PPTOD (20.94 ms), it delivers substantial improvements in both dialogue state tracking and task-oriented dialogue modeling.
Specifically, our \ours{} yields absolute gains of 5.21\% in Joint Goal Accuracy (JGA) and 3.53 points in the combined score.
This result highlights that our proposed framework strikes an effective balance between achieving state-of-the-art performance and maintaining practical runtime efficiency.

\subsection{\sechl{Case study}}
\label{sec:case_study}

To provide deeper qualitative insights into the characteristics, strengths, and weaknesses of our proposed framework, we present a detailed case study across two primary tasks: language understanding (specifically intent classification) and end-to-end (E2E) dialogue modeling.

\subsubsection{Analysis of language understanding failures}\label{sec:intent_cases}

\begin{table}[t]
\begin{center}
\begin{minipage}{1.0\textwidth}
    \setlength{\tabcolsep}{1mm}
    \caption{Failure cases from \ours{} on BANKING77 test set}\label{tab:intent_cases}
    \begin{tabular*}{\linewidth}{@{\extracolsep{\fill}}p{0.7\linewidth}cc@{\extracolsep{\fill}}}
        \toprule
        \textbf{Input} & \textbf{Golden} & \textbf{Prediction} \\
        \midrule
        \multicolumn{3}{@{}l@{}}{\textbf{\textit{Similar Intent Group:}} \{\texttt{card\_delivery\_estimate (cde)}, \texttt{card\_arrival (ca)}\}} \\
        \multicolumn{3}{@{}l@{}}{\quad\quad\textcolor{red}{\texttt{Error Rate: 14/80=17.50\%}}} \\
        \textbf{User:} \textit{I am waiting for my card to arrive.} & \texttt{cde} & \texttt{ca} \\
        \textbf{User:} \textit{How long will it take to arrive?} & \texttt{cde} & \texttt{ca} \\
        \textbf{User:} \textit{How long does a card delivery take?} & \texttt{ca} & \texttt{cde} \\
        \midrule
        \multicolumn{3}{@{}l@{}}{\textbf{\textit{Similar Intent Group:}} \makecell[l]{\{\texttt{compromised\_card (cc)}, \texttt{card\_payment\_not\_recognised (cpnr)},\\\quad\texttt{direct\_debit\_payment\_not\_recognised (ddpnr)}\}}} \\
        \multicolumn{3}{@{}l@{}}{\quad\quad\textcolor{red}{\texttt{Error Rate: 10/120=8.33\%}}} \\
        \textbf{User:} \textit{I see random purchases to my account, was it hacked?} & \texttt{cc} & \texttt{cpnr} \\
        \textbf{User:} \textit{Somebody used my card to make a purchase} & \texttt{cpnr} & \texttt{cc} \\
        \textbf{User:} \textit{I have an unauthorized transaction on my statement} & \texttt{cpnr} & \texttt{ddpnr} \\
        \textbf{User:} \textit{I was charged on my account that shouldn't be there.} & \texttt{ddpnr} & \texttt{cpnr} \\
        \textbf{User:} \textit{I am concerned about the security in my account and would like to make a dispute.} & \texttt{ddpnr} & \texttt{cc} \\
        \midrule
        \multicolumn{3}{@{}l@{}}{\textbf{\textit{Similar Intent Group:}} \makecell[l]{\{\texttt{declined\_card\_payment (dcp)}, \texttt{declined\_transfer (dt)},\\\quad\texttt{failed\_transfer (ft)}\}}} \\
        \multicolumn{3}{@{}l@{}}{\quad\quad\textcolor{red}{\texttt{Error Rate: 9/118=7.63\%}}} \\
        \textbf{User:} \textit{How can I fix my card, it got declined twice.} & \texttt{dt} & \texttt{dcp} \\
        \textbf{User:} \textit{I can't transfer money from my account.} & \texttt{dt} & \texttt{ft} \\
        \textbf{User:} \textit{My transfer did not go through.} & \texttt{ft} & \texttt{dt} \\
        \midrule
        \multicolumn{3}{@{}l@{}}{\textbf{\textit{Similar Intent Group:}} \{\texttt{topping\_up\_by\_card (tubc)}, \texttt{transfer\_into\_account (tic)}\}} \\
        \multicolumn{3}{@{}l@{}}{\quad\quad\textcolor{red}{\texttt{Error Rate: 4/79=5.06\%}}} \\
        \textbf{User:} \textit{How can someone add money to my account?} & \texttt{tubc} & \texttt{tic} \\
        \textbf{User:} \textit{Can I add funds to the card directly from my bank account?} & \texttt{tic} & \texttt{tubc} \\
        \textbf{User:} \textit{How do I top up my card?} & \texttt{tic} & \texttt{tubc} \\
        \bottomrule
    \end{tabular*}
\footnotetext{The reported error rates exclusively account for classification errors within intent groups.}
\end{minipage}
\end{center}
\end{table}

This subsection investigates the classification errors generated by \ours{}, with its backbone being the \llamaSmall{} model, in the BANKING77 test set.
As illustrated by the failure cases in Table~\ref{tab:intent_cases}, these errors mainly occur between semantically similar intents, such as \texttt{card\_delivery\_estimate} and \texttt{card\_arrival}.

The primary challenge stems from the inherent design of the original intent schema: the reliance on abbreviated intent labels that lack explicit natural language descriptions.
This deficiency introduces ambiguity that makes it difficult for the model to differentiate between similar intents, essentially forcing the model to rely excessively on its internal memorization capabilities rather than generalizable understanding.
This characteristic of these datasets favors discriminative and classification-based methods but hinders effective generalization when training on mixed datasets.
Ultimately, it restricts the flexible and effective scaling of our framework to unseen intent labels.

These specific failure cases explain the results reported in Table~\ref{table:intent_full}, where the \llamaSmall{} (FT) model fine-tuned solely on a single dataset achieves scores equal to or higher than the multi-dataset learning paradigm employed by \ours{}.
In other words, \ours{} struggles to summarize common semantic features among diverse datasets and learn generalizable understanding.
We posit that integrating explicit natural language descriptions and rich illustrative examples directly into the schema definition can effectively resolve the ambiguity between similar intents, thereby significantly enhancing the generalization of our framework in language understanding.

\subsubsection{Analysis of end-to-end dialogue modeling superiority}\label{sec:e2e_cases}

\begin{table}[t]
\begin{center}
\begin{minipage}{1.0\textwidth}
    \setlength{\tabcolsep}{1mm}
    \caption{Case study on E2E dialog modeling from MultiWOZ 2.1 test set}\label{tab:e2e_cases}
    % p{0.7\linewidth}
    \begin{tabular*}{\linewidth}
    {@{\extracolsep{\fill}}p{0.08\linewidth}p{0.9\linewidth}@{\extracolsep{\fill}}}
        \toprule
        \multicolumn{2}{@{}c@{}}{\textbf{\textit{Error Type: Incorrect Dialogue State}}} \\
        \multicolumn{2}{@{}l@{}}{\footnotesize\textbf{Dialogue: MUL2060}} \\
        \footnotesize\textbf{Turn 4} & \textbf{User:} \textit{Great I also need to find a place to stay that doesn't need to have free parking and is in the cheap price range.} \\
        \footnotesize\textbf{Oracle} & \footnotesize\textcolor{cyan}{[State] hotel: \{\textcolor{green}{parking=dontcare}, pricerange=cheap\}} \textcolor{blue}{[DB] hotel: 10} \\
        \footnotesize\textbf{Ours} &  \footnotesize\textcolor{cyan}{[State] hotel: \{\textcolor{green}{parking=dontcare}, pricerange=cheap\}} \textcolor{blue}{[DB] hotel: 10} \\
        \footnotesize\textbf{pptod} & \footnotesize\textcolor{cyan}{[State] hotel: \{\textcolor{red}{parking=no}, pricerange=cheap\}} \textcolor{blue}{[DB] hotel: \textcolor{red}{0}} \\
        \cdashline{1-2}[1.5pt/3pt] \\ [-2ex]
        \multicolumn{2}{@{}l@{}}{\footnotesize\textbf{Dialogue: PMUL1920}} \\
        \footnotesize\textbf{Turn 5} & \textbf{User:} \textit{it should be in the town center} \footnotesize\textcolor{cyan}{[State] ...} \textcolor{blue}{[DB] hotel: 0} [Delex] ... \\
        \footnotesize\textbf{Turn 6} & \textbf{User:} \textit{ok then i will go ahead and take any of the moderately priced ones with free wifi. please recommend one. any area is fine.} \\
        \footnotesize\textbf{Oracle} & \footnotesize\textcolor{cyan}{[State] hotel: \{\textcolor{green}{area=dontcare}, internet=yes, pricerange=moderate\}} \textcolor{blue}{[DB] hotel: 12} \\
        \footnotesize\textbf{Ours} &  \footnotesize\textcolor{cyan}{[State] hotel: \{\textcolor{green}{area=dontcare}, internet=yes, pricerange=moderate\}} \textcolor{blue}{[DB] hotel: 12} \\
        \footnotesize\textbf{pptod} & \footnotesize\textcolor{cyan}{[State] hotel: \{\textcolor{red}{area=centre}, internet=yes, pricerange=moderate\}} \textcolor{blue}{[DB] hotel: \textcolor{red}{0}} \\
        \midrule
        \multicolumn{2}{@{}c@{}}{\textbf{\textit{Error Type: Inappropriate Policy Planning}}} \\
        \multicolumn{2}{@{}l@{}}{\footnotesize\textbf{Dialogue: MUL0947}} \\
        \footnotesize\textbf{Turn 2} & \textbf{User:}  \textit{Perhaps the center of town. Can you provide me with information regarding the entrance fee and a telephone number too?} \footnotesize\textcolor{cyan}{[State] attraction: \{area=centre, type=museum\}} \textcolor{blue}{[DB] attraction: 11} \\
        \footnotesize\textbf{Oracle} & \footnotesize[Delex] The \textcolor{green}{[value\_name]} is located in the [value\_area] is \textcolor{green}{[value\_price]} to enter and the phone number is \textcolor{green}{[value\_phone]}. \\
        \footnotesize\textbf{Ours} & \footnotesize[Delex] \textcolor{green}{[value\_name]} is \textcolor{green}{[value\_price]} and their phone number is \textcolor{green}{[value\_phone]}. \\
        \footnotesize\textbf{pptod} & \footnotesize[Delex] \textcolor{red}{There are [value\_choice] [value\_type] in the [value\_area].} \\
        \cdashline{1-2}[1.5pt/3pt] \\ [-2ex]
        \multicolumn{2}{@{}l@{}}{\footnotesize\textbf{Dialogue: MUL1376}} \\
        \footnotesize\textbf{Turn 4} & \textbf{User:} \textit{I would be interested in a cheap guesthouse if that is available. Will you check for me?} \footnotesize\textcolor{cyan}{[State] hotel: \{pricerange=cheap, stars=4, type=guest house\}} \textcolor{blue}{[DB] hotel: 7} [Delex] There are [value\_choice] available that match what you're looking for, do you have any preference for what area you want to stay in? \\
        \footnotesize\textbf{Turn 5} & \textbf{User:} \textit{I need free internet and I need their postcode.} \footnotesize\textcolor{cyan}{[State] hotel: \{internet=yes, pricerange=cheap, stars=4, type=guest house\}} \textcolor{blue}{[DB] hotel: 7} \\
        \footnotesize\textbf{Oracle} & \footnotesize[Delex] \textcolor{green}{[value\_name]} has internet. It is located in \textcolor{green}{[value\_postcode]} postal code. \\
        \footnotesize\textbf{Ours} & \footnotesize[Delex] \textcolor{green}{[value\_name]} is available. The postcode is \textcolor{green}{[value\_postcode]}. \\
        \footnotesize\textbf{pptod} & \footnotesize[Delex] \textcolor{red}{There are [value\_choice] [value\_type] that meet your needs. Is there a particular area of town you would like to stay in?} \\
        \midrule
        \multicolumn{2}{@{}c@{}}{\textbf{\textit{Error Type: Missing Requested Information}}} \\
        \multicolumn{2}{@{}l@{}}{\footnotesize\textbf{Dialogue: MUL2405}} \\
        \footnotesize\textbf{Turn 1} & \textbf{User:} \textit{Hi. I am having trouble finding a place to eat called prezzo. Can you locate it for me and tell me more?} \footnotesize\textcolor{cyan}{[State] restaurant: \{name=prezzo\}} \textcolor{blue}{[DB] restaurant: 1} \\
        \footnotesize\textbf{Oracle} & \footnotesize[Delex] [value\_name] is located at \textcolor{green}{[value\_address]}, in the [value\_area]. It's an [value\_food] restaurant. [value\_price] price range. \\
        \footnotesize\textbf{Ours} & \footnotesize[Delex] [value\_name] is an [value\_price] [value\_food] restaurant in the [value\_area] part of town. The address is \textcolor{green}{[value\_address]} and the phone number is [value\_phone]. \\
        \footnotesize\textbf{pptod} & \footnotesize[Delex] \textcolor{red}{[value\_name] is an [value\_price] [value\_food] restaurant in the [value\_area].} \\
        \bottomrule
    \end{tabular*}
% \footnotetext{Dialogue states, database results and delexicalized responses are marked in different colors.}
\footnotetext{Detailed database results and system actions are omitted for brevity.}
\end{minipage}
\end{center}
\end{table}

This analysis focuses on the E2E dialogue modeling capabilities of \ours{} in comparison to the PPTOD baseline, both of which are built on the Qwen2.5-3B-Instruct model.
Table~\ref{tab:e2e_cases} presents comparative cases on the MultiWOZ 2.1 test set

\bmhead{Error Type: Incorrect Dialogue State}
In this category, the baseline PPTOD model exhibits significant errors in dialogue understanding, resulting in no matching items in subsequent database queries.
\begin{itemize}
    \item Dialogue MUL2060 (Turn 4): The user explicitly requests ``\textit{a place that doesn't need to have free parking}'', implying \texttt{parking=dontcare} as it's not a hard constraint. However, PPTOD incorrectly updates the dialogue state to \texttt{parking=no}.
    \item Dialogue PMUL1920 (Turn 5): PPTOD fails to accurately track the user's preference shift, retaining the \texttt{area=centre} constraint despite the user indicating an intention to accept ``\textit{any area is fine}'', which should result in \texttt{area=dontcare}.
\end{itemize}

In contrast, our \ours{} model correctly predicts the desired dialogue states throughout both dialogues, demonstrating superior state tracking competence which is essential for successful task completion.

\bmhead{Error Type: Inappropriate Policy Planning}
The failure of PPTOD in `MUL1376' underscores a clear advantage for \ours{} in strategic decision-making.
Both methods initially choose to inquire about the user's \texttt{area} preference in Turn 4.
However, in Turn 5, PPTOD executes a redundant dialogue act, leading to an inappropriate response and a failed interaction.
Conversely, \ours{} promptly pivots its policy toward recommending an available hotel in the subsequent turn, enabling a smooth task completion.
This validates the superior ability to plan policies embedded within our framework.

\bmhead{Error Type: Missing Requested Information}
A third common failure mode for PPTOD involves omission of critical information.
In response to the user's request in Turn 1 of `MUL2405', PPTOD neglects to provide the requested \texttt{address} information concerning the restaurant.
\ours{}, however, retrieves and presents all necessary entity attributes, ensuring task success and complete user satisfaction.

Collectively, these cases validate the superior dialogue understanding, state tracking, and policy planning abilities of \ours{} compared to the baseline method.
\section{\sechl{Limitations and future directions}}\label{sec:limitations}

In this section, we discuss the limitations of our proposed framework and outline promising directions for future research.

\bmhead{Design trade-offs in modeling language understanding}
Natural Language Understanding (NLU), encompassing tasks like intent detection and dialogue state tracking, is a cornerstone of task-oriented dialogue (TOD) systems.
Our framework models these tasks within a schema-aware generative paradigm.
Compared to non-scalable, classification-based approaches that are tailored for specific tasks, this generative formulation may not achieve the same peak performance on certain NLU benchmarks.
However, the primary advantage of our approach lies in its flexibility and scalability; the schema-aware generative model can be extended to known or unseen scenarios with remarkable adaptability, often without the need for complete retraining.

\bmhead{Framework complexity vs. Simple LLM fine-tuning}
A pertinent question is whether simply fine-tuning a large language model (LLM) for each specific scenario offers a better trade-off between simplicity and performance.
Indeed, fine-tuning an LLM on a single data-rich domain is a straightforward method to achieve considerable results.
In contrast, our unified framework, \ours{}, requires more upfront efforts in curating schema definitions and organizing data from multiple scenarios.

While this may yield only modest performance gains in single-domain settings, its strength becomes evident in practical applications characterized by diverse dialogue scenarios.
The alignment mechanisms in \ours{} are designed not only to mitigate interference between different scenarios but also to facilitate knowledge transfer and utilization among similar ones.
In other words, our framework demonstrates superior generalization in low-resource settings.
In summary, while direct fine-tuning is suitable for isolated, data-abundant tasks, \ours{} presents a more balanced and robust solution for the majority of real-world applications involving multiple, varied, or low-resource domains.

\bmhead{Scalability constraints}
As discussed in Section~\ref{sec:low_resource}, \ours{} significantly outperforms baseline methods in low-resource and unseen scenarios, demonstrating strong generalization capabilities in end-to-end dialogue modeling.
The proposed alignment mechanisms contribute to this enhanced extensibility to unseen domains without retraining.
Nevertheless, there remains room for improvement.
For instance, as shown in Table~\ref{table:zero_shot}, the performance on certain tasks in unseen domains, such as the 26.9\% joint goal accuracy for dialogue state tracking, has not yet reached the desired level.
To address the performance gaps in unseen domains, a primary focus of our future work will be on data-centric approaches.
We believe that incorporating more diverse datasets and developing sophisticated data augmentation techniques can substantially improve the model's robustness and generalization capabilities.

Additionally, our framework's ability to generalize is constrained when faced with unseen intent labels that are highly abbreviated and lack clear semantic definitions. 
The absence of natural language explanations for these labels forces the model to rely heavily on memorization, hindering its flexibility and effectiveness.
% Building on the limitation regarding ambiguous intent labels.
To address this, future improvements could involve exploring the integration of rich, explicit natural language descriptions into the schema.
Such an enhancement could resolve ambiguity among similar intents and reduce the model's reliance on memorization.
Ultimately, this may enable our framework to effectively unify a wide range of intent classification datasets, paving the way for a more universal dialogue model.

\bmhead{Real-world applicability assessment}
The high cost of annotation means that most real-world TOD datasets are limited in scale and scenario diversity, which in turn constrains the performance evaluation of methods in practical applications.
Although some recent work has used LLMs to synthesize training data in various domains~\citep{li-etal-2022-controllable,niu-etal-2024-enhancing}, a significant quality gap persists between the synthetic and real data.
Consequently, synthetic data is typically used only for the initial pre-training phase, followed by fine-tuning on real data.
Therefore, synthesizing high-quality, diverse data to improve or dynamically evaluate TOD models remains a promising avenue for future work.

Equally important is the development of more flexible and fine-grained evaluation methodologies~\citep{DBLP:journals/corr/abs-2504-19982,DBLP:journals/corr/abs-2505-05445}.
Current automatic metrics often fail to capture critical intermediate errors that can occur during multi-turn user-agent interactions.
Future work should aim to design novel evaluation techniques that complement traditional metrics, providing a more holistic and realistic assessment of the applicability of a dialogue system in the real world.

\section{Related work}\label{sec:related_work}

\subsection{Pre-trained conversation models (PCMs)}
Transformer-based~\citep{DBLP:conf/nips/VaswaniSPUJGKP17} pre-trained language models (PLMs)~\citep{DBLP:conf/naacl/DevlinCLT19,DBLP:conf/nips/YangDYCSL19,DBLP:journals/jmlr/RaffelSRLNMZLL20,DBLP:conf/nips/00040WWLWGZH19}, which master the deep contextualized understanding of natural language texts, have garnered increasing attention and achieved remarkable success in a variety of natural language processing (NLP) tasks, such as text classification, sequence labeling, and text generation.
However, many prior studies have highlighted the intrinsic differences in linguistic patterns between human conversations and the general texts used for pre-training PLMs~\citep{DBLP:conf/acl/ZhangSGCBGGLD20}.
PLMs directly fine-tuned on downstream dialog tasks often struggle to capture conversational linguistic features, failing to achieve the significant improvements observed in other NLP tasks.
To address this issue, further pre-training PLMs on large-scale dialog corpora with well-designed dialog objectives has been proposed to build pre-trained conversation models (PCMs) that enrich conversational knowledge~\citep{DBLP:journals/corr/abs-2009-13570,DBLP:conf/acl/ZhangSGCBGGLD20,DBLP:conf/sigir/HeDYSHSL22}.
These PCMs offer superior initialization parameters for fine-tuning on downstream dialog tasks and can be broadly classified into two categories: those focusing on open-domain conversations and those tailored for task-oriented dialogs.

DialoGPT~\citep{DBLP:conf/acl/ZhangSGCBGGLD20} is a representative PCM based on the GPT-2~\citep{radford2019language} architecture, pre-trained on a large-scale dialog corpus collected from Reddit comment chains to enhance the relevance, informativeness, and fluency of open-domain dialog responses.
Blender~\citep{DBLP:conf/eacl/RollerDGJWLXOSB21} extended the scale of open-domain dialog data for pre-training, demonstrating powerful response generation capabilities.
Furthermore, additional pre-training tasks have been employed to improve performance~\citep{DBLP:conf/emnlp/HendersonCMSWV20}, such as latent act recognition in PLATO~\citep{DBLP:conf/acl/BaoHWWW20} and response selection.

Another majority of PCMs are dedicated to tackling various challenges in task-oriented dialog scenarios, such as multi-turn dialog understanding, dialog state tracking, and dialog policy planning.
These PCMs can be further categorized into three types.
The first type of PCMs aims to learn semantic representations that improve dialogue understanding performance by pre-training on conversational data without relying on any additional human annotations~\citep{DBLP:conf/acl/ZengHWZWXX23}.
For instance, ConvBERT~\citep{DBLP:journals/corr/abs-2009-13570} further trained BERT using a masked language modeling objective, while TOD-BERT~\citep{DBLP:conf/emnlp/WuHSX20} additionally incorporated a response selection objective.
DialogueBERT~\citep{DBLP:conf/cikm/ZhangGC21} replaced the conventional masked language modeling in TOD-BERT with a masked utterance modeling objective.
The second type defines pre-training objectives based on the annotations of TOD subtasks, empowering PLMs with the necessary skills required to complete specific dialogue tasks.
GALAXY~\citep{DBLP:conf/aaai/HeDZWCLJYHSSL22} explicitly captured dialog policy in labeled and unlabeled dialogs through semi-supervised learning of dialog act prediction task.
SPACE-2~\citep{DBLP:conf/coling/HeDHYCDHSL22} transformed the semantic frame of each dialog turn into a semantic tree structure (STS) and established a contrastive learning objective based on semantic tree similarity.
The pre-trained SPACE-2 achieved state-of-the-art performance across multiple dialog understanding tasks, including intent prediction, slot filling, semantic parsing, and dialog state tracking.
PPTOD~\citep{DBLP:conf/acl/SuSMG0LZ22} conducted multi-task dialog pre-training using labeled data from various TOD tasks.
The third type involves constructing PCMs that explicitly or implicitly model the task flows inherent in task-oriented dialog data in an end-to-end manner.
SOLOIST~\citep{DBLP:journals/tacl/PengLLSLG21} fine-tuned the GPT-2 model on two TOD datasets to sequentially generate the results of multiple TOD subtasks.
SPACE-3~\citep{DBLP:conf/nips/00040WWLWGZH19} maintained a task flow consisting of four modules within a single transformer and completed semi-supervised end-to-end pre-training by leveraging only the semantic frame as an intermediate task annotation.
The latter two types of methods bridge the gap between pre-training and downstream fine-tuning by directly modeling the relationship between dialog sessions and various task annotations.

\subsection{Large Language Models (LLMs)}
The transition from PLMs to LLMs was characterized by a substantial increase in both model size and the scale of training data.
Pre-training billions of parameters on trillions of tokens spanning diverse text genres (e.g., news, books, Wikipedia articles, programming codes) allows LLMs to better comprehend the context and produce more coherent content through a simple next-token prediction objective, such as \llama{} series~\citep{DBLP:journals/corr/abs-2307-09288}, GPT-3 series~\citep{DBLP:conf/nips/BrownMRSKDNSSAA20,DBLP:conf/nips/Ouyang0JAWMZASR22}, GLM series~\citep{DBLP:conf/acl/DuQLDQY022,DBLP:conf/iclr/ZengLDWL0YXZXTM23}, \hlyellow{Qwen series~\citep{DBLP:journals/corr/abs-2412-15115}}, and so on.
LLMs have revolutionized the implementation of various NLP applications, achieving remarkable success across a range of NLP tasks while demonstrating impressive few-shot and even zero-shot generalization capabilities.
Instruction fine-tuning (IFT)~\citep{DBLP:conf/iclr/WeiBZGYLDDL22}, which involves fine-tuning a pre-trained LLM on a set of (instruction, input, output) triples, further enhances LLMs' fine-grained understanding and execution capabilities for complex instructions, thus broadening their applicability to new tasks and scenarios.
\hlyellow{IFT has motivated researchers to reconsider TOD architectures.
For instance, the AutoTOD~\citep{DBLP:conf/acl/XuMYSH24} approach proposes a shift from complex modularity to a zero-shot autonomous agent.
This model requires only a general-purpose instruction-following LLM (e.g., GPT-4) and simple task descriptions to autonomously decide all dialogue actions.
Spec-TOD~\citep{DBLP:journals/corr/abs-2507-04841} incorporates explicit task instructions into the training framework to obtain a specialized instruction-tuned LLM.
Spec-TOD also focuses on reducing the reliance on massive amounts of labeled data, particularly in low-resource scenarios.}

Our work is inspired by the success of IFT on LLMs and aims to explore the potential of empowering LLMs with superior performance in handling task-oriented dialog (TOD) workflows under regular and low-resource settings.
Benefiting from the extraordinary text modeling ability of LLMs, compared with PCMs, LLMs have advantages in reducing the degree of catastrophic forgetting and generating structured contents for building TOD systems.
Specifically, continually pre-training and then fine-tuning PLMs on dialog data leads to increased catastrophic forgetting, which limits their generalization performance on downstream dialog tasks.
In contrast, directly performing IFT on LLMs can leverage annotated information from various TOD datasets to build an efficient and generalizable task-oriented dialog system while preserving much of the general knowledge acquired during pre-training.
Additionally, LLMs excel at converting natural texts into structured forms (e.g., dictionary and list objects in Python\footnote{\url{https://www.python.org/}}), which aids in generating and parsing the results for multi-intent recognition, dialog state tracking, and other TOD tasks.
This capability simplifies the design and implementation of TOD systems by facilitating the structured output crucial for these tasks.

\hlyellow{As LLMs increase the sophistication of TOD systems, traditional evaluation metrics prove insufficient.
The TD-EVAL framework~\citep{DBLP:journals/corr/abs-2504-19982} unifies evaluation by combining fine-grained turn-level analysis with holistic dialogue-level comparisons, demonstrating excellent alignment with human judgment.
Furthermore, \citet{DBLP:conf/acl/BaidyaDG25} proposes a framework to quantify and analyze the discrepancy between AI agent and human expert behavior in complex TOD systems.
This work identifies that the performance gap in zero-shot LLM agents is significantly correlated with the widening of the behavior gap, including discrepancies in dialog acts, tool usage, and knowledge utilization.
Increasing task complexity highlights the need for improved behavioral alignment strategies.}
\section{Conclusion}
% In this work, we introduce \ours{}, an innovative instruction-aware and schema-aware end-to-end task-oriented dialog (TOD) system, which could seamlessly adapt to a variety of annotation schemas and execute diverse TOD workflows within a unified framework.
In this work, we introduce \ours{}, an innovative instruction-tuning framework for end-to-end task-oriented dialog (TOD) modeling.
\hlyellow{Our approach moves beyond simply fine-tuning a powerful LLM backbone by introducing a structured methodology that enables a single model to seamlessly adapt to a variety of annotation schemas and execute diverse TOD workflows within a unified framework.}
To facilitate effective alignments of task instructions with corresponding workflows as well as dialogue schemas with task outputs, we integrate two novel alignment mechanisms during the construction and modeling of the TOD instruction fine-tuning data.
Meanwhile, our framework manages dialogue interactions in a session-level end-to-end manner, where tasks such as language understanding, policy planning, and response generation are sequentially executed for each turn, conditioned on both the dialog history and the outcomes of previous task flows.
% Experimental results on multiple public benchmarks demonstrate that \ours{} significantly enhances dialog task completion and even outperforms most advanced specialized models in dialog state tracking, which verifies the efficacy of our proposed instruction fine-tuning architecture in improving end-to-end task-oriented dialog modeling.
% Further analysis indicates that our schema-aware dialog modeling not only bolsters the system's generalization capabilities and data utilization efficiency in low-resource settings but also strengthens its robustness against data noise and error propagation in multi-turn dialog interactions.
Our experimental results demonstrate that this approach is highly effective.
\hlyellow{While achieving competitive performance against specialized models, our primary contribution is not focused on state-of-the-art metrics alone.
Instead, we have shown that the true strength of \textsc{ESAinsTOD} lies in its ability to systematically leverage the power of LLMs for complex real-world TOD challenges.
Specifically, our in-depth analysis confirms that the proposed framework significantly enhances generalization capabilities, improves data utilization efficiency in low-resource settings, and strengthens robustness against error propagation in multi-turn dialog interactions.
By structuring the learning process around schemas and instructions, our work provides a new perspective for building robust and adaptable TOD systems in the era of LLMs.}

\backmatter

\section*{Acknowledgments}
We thank all anonymous reviewers for their constructive comments.
This work was supported by the National Natural Science Foundation of China (NSFC) via grant 62306342, 62236004, 62206078 and 62441603, and Du Xiaoman (Beijing) Science Technology Co., Ltd.

\section*{Data availability}
All datasets listed in Table~\ref{table:datasets} are publicly available.
SGD: \url{https://github.com/google-research-datasets/dstc8-schema-guided-dialogue}.
Frames dataset is available at \url{https://github.com/awslabs/pptod}.
BiToD: \url{https://github.com/HLTCHKUST/BiToD}.
STAR: \url{https://github.com/RasaHQ/STAR}.
SNIPS: \url{https://github.com/LeePleased/StackPropagation-SLU}.
Other datasets can be downloaded from \url{https://github.com/AlibabaResearch/DAMO-ConvAI/tree/main/space-3}.

\newpage
\clearpage

\bibliography{sn-bibliography}% common bib file


\end{document}